\documentclass[10pt,twocolumn,letterpaper]{article}

\usepackage[pagenumbers]{cvpr} % To force page numbers, e.g. for an arXiv version

\definecolor{cvprblue}{rgb}{0.21,0.49,0.74}
\usepackage[pagebackref,breaklinks,colorlinks,allcolors=cvprblue]{hyperref}

\usepackage{array} % For better column definitions
\usepackage{booktabs} % For professional-looking tables (\toprule, \midrule, \bottomrule)
\usepackage{bm}
\usepackage{comment}
\usepackage{multirow}
\usepackage{multicol}
\usepackage[capitalize]{cleveref}
\usepackage{graphicx}
\crefname{figure}{Fig.}{Figs.}
\Crefname{figure}{Fig.}{Figs.}

\newcommand{\themethod}{Muskie\xspace}
\newcommand{\lsd}[1]{{\color{orange}[\textbf{LSD:} #1]}\xspace}
\usepackage{soul}
\newcommand{\lsddel}[1]{{\color{orange}\st{#1}\xspace}}

\makeatletter
\renewcommand{\paragraph}{%
    \@startsection{paragraph}{4}%
    {\z@}{-0.5em}{-0.5em}%
    {\normalfont\normalsize\bfseries}%
}
\makeatother

\definecolor{cvprblue}{rgb}{0.21,0.49,0.74}
\usepackage[pagebackref,breaklinks,colorlinks,allcolors=cvprblue]{hyperref}

\def\paperID{{***}} % *** Enter the Paper ID here
\def\confName{CVPR}
\def\confYear{2026}

\title{Muskie: \underline{Mu}lti-view Ma\underline{sk}ed \underline{I}mag\underline{e} Modeling for 3D Vision Pre-training}

\author{
Wenyu Li$^{\dagger}$
\and 
Sidun Liu$^{\dagger}$
\and 
Peng Qiao$^*$
\and
Yong Dou$^*$   
\and
Tongrui Hu
\and
\\
National University of Defense Technology
}

\begin{document}

\twocolumn[{
\maketitle

\begin{center}
\includegraphics[width=0.98\linewidth]{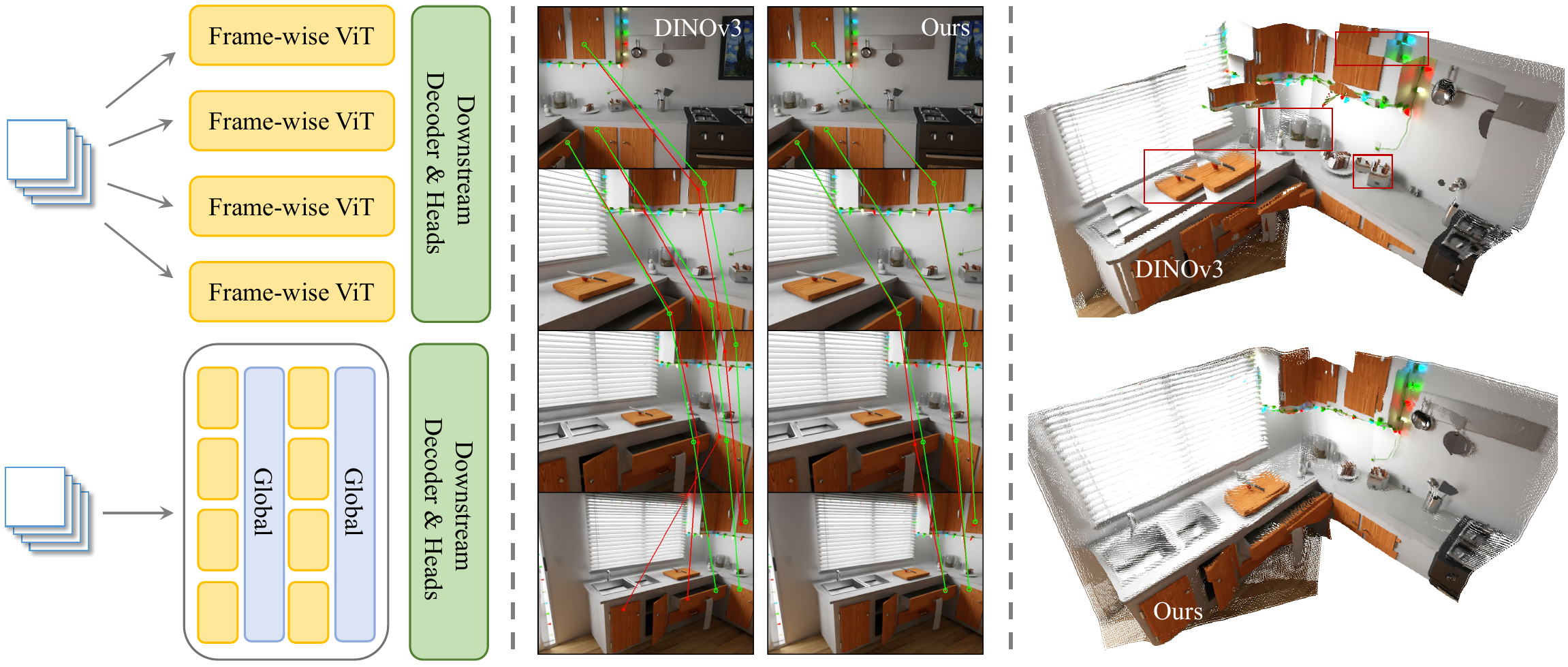}
\end{center}
\vspace{-0.5cm}
\captionsetup{type=figure}
\captionof{figure}
{
% \textbf{\themethod}
% is a  vision model with 3D awareness trained without 3D supervision and process a dynamic number of images at once.
% Our approach leverages the idea of inpainting and adopt it for multi-view images. 
% \themethod is trained to find and utilize geometric correspondences from other views to perform reconstruction.
\textbf{\themethod} is a native multi-view visual backbone designed for 3D vision tasks. 
Through Multi-view Masked Image Modeling pre-training, it learns to jointly extract representations across multiple views in a single forward pass.
This is in contrast to the conventional frame-wise paradigm, where a ViT independently encodes each view before features are fused. 
\themethod establishes stronger multi-view consistency, demonstrated by predicted tracks~(red) that align closely with the ground truth~(green).
Using \themethod as a backbone also leads to superior performance in applications like pointmap estimation, where it produces coherent and geometrically complete 3D reconstructions.
}

\label{fig:teaser}
\vspace{0.4cm}
}]

\renewcommand{\thefootnote}{\fnsymbol{footnote}}
\footnotetext{$^\dagger$Equal Contribution\hspace{0.2cm} $^*$Corresponding Author}

\begin{abstract}
We present \themethod, a native multi-view vision backbone designed for 3D vision tasks.
Unlike existing models, which are frame-wise and exhibit limited multi-view consistency, \themethod is designed to process multiple views simultaneously and introduce multi-view consistency in pre-training stage.
\themethod is trained to reconstruct heavily masked content in one view by finding and utilizing geometric correspondences from other views. 
Through this pretext task and our proposed aggressive masking strategy, the model implicitly to learn view-invariant features and develop strong geometric understanding without any 3D supervision.
Compared with state-of-the-art frame-wise backbones such as DINO, \themethod achieves higher multi-view correspondence accuracy. Furthermore, we demonstrate that using \themethod as a backbone  consistently enhances performance on downstream 3D tasks, including camera pose estimation and pointmap reconstruction.
Codes are publicly available at \href{https://leo-frank.github.io/Muskie/}{https://leo-frank.github.io/Muskie/}. 
\end{abstract}    
\section{Introduction}
\label{sec:intro}

Reconstructing and understanding the 3D world from 2D images is one of the long-standing and fundamental goals in computer vision~\cite{triggs00bundle, triggs99camera, schonberger16structure-from-motion,schoenberger2016mvs}. 
A prevailing practice in this domain is to leverage Vision Foundation Models~(VFMs), such as the DINO series~\cite{oquab24dinov2:,simeoni2025dinov3}, as powerful feature extractors for downstream 3D tasks like 3D reconstruction and camera pose estimation.
These VFMs, pre-trained on massive-scale 2D image datasets, have achieved remarkable success and become the de-facto backbones for many modern approaches~\cite{wang2025vggt, streamVGGT, wang2025pi3, keetha2025mapanything, deng2025vggtlongchunkitloop}.

However, recent studies~\cite{ICCV21prior3D, probe3d} revealed that existing VFMs exhibit limited 3D understanding, notably a lack of \textit{multi-view consistency}, which requires the feature representations to be consistent across different viewpoints, as shown in ~\cref{fig:teaser}.
% In particular, as illustrated in \cref{fig:teaser}, they struggle with the property of \textit{multi-view consistency}, which requires the feature representations to be consistent for different viewpoints of the scene.
This capability is a core requirement for 3D tasks as it enables the accurate aggregation of information across views.
% ~\cite{schoenberger2016mvs,schoenberger2016sfm,tang14co-localization}.
% reconstruction\cite{schoenberger2016mvs,schoenberger2016sfm} and localization pipelines\cite{tang14co-localization}.
While prior efforts have sought to instill 3D awareness into VFMs by leveraging supervisory signals from depth maps~\cite{ICCV21prior3D} or camera poses~\cite{wangqianqian2020learning}, their reliance on 3D annotations limits scalability and generalizability.
% which inherently limits their understanding of the underlying 3D structure of our world. 

\begin{comment}
\lsddel{A key capability for a model to be truly 3D-aware is \textit{multi-view consistency} :  its}\lsd{In particular, they lack the ability to relate observations of the same scene from different viewpoints—a property essential for geometric reasoning.
This property, referred to as \textit{multi-view consistency} requires that} feature representations should remain invariant to viewpoint changes, thereby allowing for the establishment of accurate correspondence.
Recent work has demonstrated that even state-of-the-art VFMs struggle with this property\cite{probe3d}.
This limitation, as illustrated in \cref{fig:teaser}, represents a significant bottleneck for progress in 3D vision.
\end{comment}

\begin{comment}
We posit that the cause for the poor multi-view consistency observed in existing VFMs lies in their frame-wise pretraining paradigm, which operates on single-view images and thus is inherently limited in capturing the relationships across views.
Consequently, in multi-view settings, these models process each image independently and fail to capture consistency across views.
Our key insight is that multi-view consistency can be fundamentally improved by leveraging multi-view data during pre-training.
\end{comment}

We posit that the key to achieving robust multi-view consistency lies in addressing it directly during the self-supervised pre-training phase.
% rather than treating it as a downstream adaptation problem.
This insight stems from a core limitation in existing VFMs: their frame-wise \mbox{pre-training} paradigm. This paradigm operates only on single-view images, providing no incentive for the model to learn cross-view relationships. Consequently, when these models are applied to multi-view settings, they process each image independently and fail to establish consistency.
% This challenge motivates our work: \textit{How can we design a pre-training scheme that explicitly utilizes  multi-view images to enhance multi-view consistency?}

\begin{comment}
To this end, we present \themethod, a native multi-view self-supervised framework that extends Masked Image Modeling (MIM) to the multi-view domain.
Unlike conventional single-view MIM that reconstruct masked regions from visible parts based on semantic priors, \themethod reconstructs them by discovering and leveraging multi-view geometric correspondences.
Through this process, the model internalizes geometric priors from multi-view observations, leading to view-consistent features without explicit 3D supervision.
\end{comment}

To this end, we propose \themethod to instill multi-view consistency during pre-training by introducing the multi-view completion task, which extends Masked Image Modeling (MIM)\cite{he21masked,xie2022simmim} to the multi-view domain. 
Given multi-view observations as input, the core idea of this pretext task is to reconstruct masked regions in one view by leveraging the visible content from other available views. 
This pretext task implicitly compels the model to discover geometric correspondences, resulting in view-consistent features without explicit 3D annotations.

To make the pre-training scheme effective, we first need to prevent the model from taking shortcuts—that is, reconstructing masked regions using only intra-view cues like conventional MIM methods.
We address this by proposing an aggressive masking strategy with high masking ratios and spatially concentrated mask blocks.
By masking large and contiguous blocks, this method makes single-view reconstruction nearly impossible, thus compelling the model to seek correspondences across views.
Furthermore, to  process multi-view inputs, we incorporate Alternating Attention\cite{wang2025vggt}, which facilitates efficient information exchange both intra-view and inter-view.

\begin{comment}
\themethod is trained on a challenging pretext task: it must reconstruct heavily masked content in one view by explicitly finding and leveraging geometric correspondences from other available views. 
Crucially, this is achieved without any form of 3D supervision, such as depth maps or camera poses.

\lsd{To this end, we present Muskie, a native multi-view self-supervised framework that extends Masked Image Modeling (MIM) to the multi-view domain.
Unlike conventional single-view MIM methods—which reconstruct masked content using unmasked regions and semantic understanding, thereby injecting semantic priors—Muskie reconstructs masked regions by discovering geometric correspondences and modeling cross-view transformations among multiple views.
Through this process, the model acquires geometric inductive biases, leading to view-consistent feature representations, all learned without any explicit 3D supervision, such as depth maps or camera poses.
\end{comment}

\begin{comment}
To make pre-training scheme effective, we incorporate two key design choices.
First, to enable the model to natively process multi-view inputs, our network architecture adopts Alternating Attention, which facilitates efficient information exchange both intra-view and inter-view.
Second, we adopt an aggressive masking strategy with high masking ratios and spatially concentrated mask blocks. This is crucial to prevent the model from taking shortcuts by relying solely on single-view cues for reconstruction.
\end{comment}

We pre-train \themethod on large-scale, unlabeled multi-view image collections.
The trained model demonstrates superior multi-view correspondence accuracy over state-of-the-art visual backbones, such as DINO family\cite{oquab24dinov2:,simeoni2025dinov3} and MAE\cite{he21masked}.
We further evaluate it as a backbone within end-to-end 3D reconstruction frameworks~\cite{wang2025pi3} for pointmap and camera pose estimation, observing consistent improvements across these downstream tasks.
The results show that our multi-view pre-training effectively builds geometric understanding into visual features, enabling 3D-aware representations.
In summary, our contributions are three-fold: 
\begin{itemize}
    \item We introduce \themethod, a novel pre-training framework that learns 3D-aware representations from multi-view images without any 3D supervision. 
    \item We show that \themethod produces superior geometric consistency, demonstrating that our pre-training  effectively equips the model with geometric reasoning capability.
    \item We demonstrate that using \themethod as a feature extractor enhances the performance of downstream 3D tasks, outperforming state-of-the-art backbones that are restricted to single-image pre-training. 
\end{itemize}

\begin{figure*}[!htbp]
    \centering
    \includegraphics[width=\linewidth]{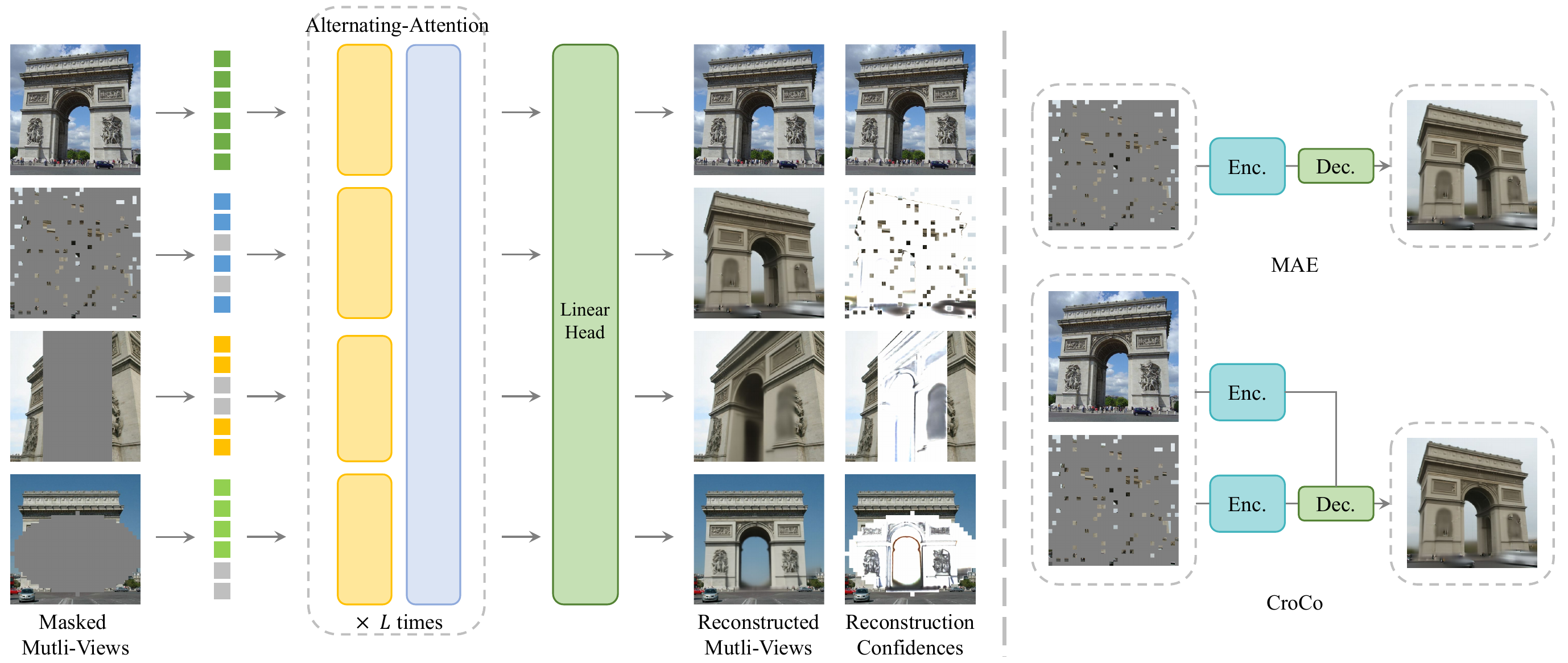}
    \caption{
    \textbf{Overview of Muskie architecture.}
    Multi-view images are divided into patches, and a portion of them is masked using various masking shapes (random, rectangular, or elliptical), replaced with learnable tokens. A subset of views is kept unmasked to serve as reference for others.
    These patches are jointly processed via stacked alternating-attention blocks\cite{wang2025vggt}.
    % \lsddel{ (global self-attention across all views and frame-wise self-attention within each view, repeated $L$ times), with a few register tokens added for conditioning and Rotary Positional Embeddings for relative patch positioning.} 
    A lightweight linear head reconstructs the masked patches along with confidence maps.
    % by leveraging geometric correspondences from unmasked regions, facilitating cross-view feature learning. 
    % Right: Comparison to single-view baselines MAE\cite{he21masked} and cross-view CroCo\cite{weinzaepfel2022croco,weinzaepfel2023croco}, highlighting M$^2$AE's multi-view processing. 
    For comparison, MAE\cite{he21masked} performs Masked Image Modeling (MIM) in a single-view setting, while CroCo\cite{weinzaepfel2022croco,weinzaepfel2023crocov2} extends MIM to dual views but still encodes each view independently during the encoding stage.}
    \label{fig:method}
    \vspace{-0.3cm}
\end{figure*}

\section{Related Works}
% Our work is broadly related to efforts to design self-supervised  vision foundation models to understand and reconstruct 3D world.

\paragraph{Self-supervised learning from 2D images.}
The rise of Vision Foundation Models~(VFMs), trained via self-supervised learning on vast 2D image datasets without annotations, has marked a paradigm shift in computer vision.
These models learn powerful general-purpose representations typically through one of two dominant approaches: contrastive learning~\cite{caron2021dino,oquab24dinov2:,simeoni2025dinov3,he19momentum,grill20bootstrap,chen2020simple}, which learns by leveraging discriminative signals between images, or masked image modeling~\cite{he21masked, tong2022videomae, xie2022simmim}, inspired by the masked token prediction task in BERT\cite{devlin18bert:}. 
Both approaches excel at learning robust dense features, such as semantic object boundaries, leading to high performance on downstream tasks such as segmentation\cite{abouzeid2025ditr, li2022maskdino} and depth estimation\cite{depth_anything_v1, depth_anything_v2, wang24moge:}.
However, these methods are trained exclusively on unstructured 2D image collections, thus lacking knowledge of the underlying 3D structure.

\paragraph{Incorporating 3D Awareness into VFMs.}
To enhance the 3D understanding of VFMs, a key research direction is to instill geometric awareness during self-supervised pre-training. 
Early approaches focused on incorporating explicit 3D priors into contrastive learning frameworks. For instance, Pri3D\cite{ICCV21prior3D} enforced multiview consistency using geometric correlations from RGB-D scans, while other works leveraged publicly available CAD models\cite{Arsomngern2023LearningGPCAD} or known relative camera poses\cite{wangqianqian2020learning}. 
However, the reliance on such annotated 3D data limit the scalability and performance gains.
A more recent line of work, CroCo\cite{weinzaepfel2022croco, weinzaepfel2023crocov2}, adapts the masked image modeling strategy to stereo data to learn 3D priors.
% , Gupta2023SiameseMA, tong2022videomae.
However, this approach is limited by its pair-wise input structure, which constrains its applicability to stereo-only tasks. 
% In contrast, our proposed \themethod introduces a scalable framework capable of processing multiple images simultaneously and maintain multi-view consistency.
% Recent studies\cite{probe3d} posit 3D awareness implies consistency of representations across multiple views and find and find existing VFMs struggle multiview consistency.
% plays a key role in understanding the 3D world.
% This capability is important because it would allow the model to correctly aggregate information across views.
% , where a second view of the same scene is added to MIM. 
% This is well suited to geometric downstream tasks.
% However, it does not scale to more views as its unscalable architecture.
% encode basic 3D properties of the surface as distances and orintations. Beyond a single image, 3D aware representations are consistent across views of the same object or scene, as they are projections of the same underlying 3D geometry\cite{el2024probing}.

% Recent self-supervised models such as DINOv2\cite{oquab24dinov2:} learn representations that encode depth and surface normals, with StableDiffusion\cite{rombach2022high_stablediffusion} [69] being a close second. Meanwhile, the training in vision language for models such as CLIP\cite{radford21learning} exhibits very poor performance despite its impressive semantic generalization capabilities.

\paragraph{Feed-Forward 3D Reconstruction Models.}
Recent learning-based approaches have shifted toward end-to-end frameworks that predict 3D structure from multi-view unposed images within a single pass\cite{wang24dust3r:, yang2025fast3r, keetha2025mapanything, streamVGGT, deng2025vggtlongchunkitloop, wang2025vggt, wang2025pi3}.
% The feed-forward transformer-based architectures  with minimum inductive biases have emerged to 
Among these works, VGGT\cite{wang2025vggt} first scales the model to a 1.2B parameter transformer that jointly predicts intrinsics, extrinsics and pointmaps.
$\pi^3$\cite{wang2025pi3} furthermore finds that the reliance on a fixed reference view can lead to failures if the reference is suboptimal and propose to predict local pointmaps without any reference frames.
% StreamVGGT\cite{streamVGGT} extends VGGT to process for efficient, real-time streaming 4D visual geometry reconstruction for on-the-fly image sequences.
% MapAnything\cite{keetha2025mapanything} take various inputs like images, camera intrinsics, poses, or depth to directly regresses the metric 3D geometry and cameras, thus addressing a broad range of 3D vision tasks in a single feed-forward pass.
% VGGT-long\cite{keetha2025mapanything} proposes chunk-based processing strategy combined with overlapping alignment and loop closure optimization to reconstruct unbounded kilometer-scale environments.
% These modern approaches all leverage DINOv2\cite{oquab24dinov2:} as image encoder because it's pretrained on a large, curated dataset and shows powerful emergent properties
% We are inspired by this line of work and we demonstrate that the reliance on DINO\cite{oquab24dinov2:} which process input images in a frame-wise manner, is suboptimal to in terms of multi-view tasks.
These modern approaches leverage the DINOv2\cite{oquab24dinov2:} as image encoder due to its curated pre-training and powerful emergent properties. However, we demonstrate that the frame-wise processing leads to poor consistency and is suboptimal for 3D tasks.
% DUSTer -> VGGT -> Pi3, streamVGGT, VGGT-long, Fast-VGGT, MapAnything (all works rely on DINOv2 as Image encoder)
% DINOv2 to be optimal in terms of downstream performance, convergence speed, and generalization

% Pioneering efforts in this domain, such as Dust3R\cite{wang24dust3r:}, focused on processing image pairs to predict a point cloud within the coordinate system of the first camera. 
% While effective for two views, scaling this to larger scenes requires a subsequent global alignment step, a process that can be both time-consuming and prone to instability.
% Subsequent work has focused on overcoming this limitation. VGGT\cite{wang2025vggt} leverages multi-task learning and large-scale datasets to achieve superior accuracy and performance
% We are inspired by this line of work, but note that it differs in objective from our analysis, as we are interested in understanding 3D awareness in models trained \textit{without} 3D supervision.

\section{Method}
\begin{figure*}[!htbp]
    \centering
    \includegraphics[width=0.97\linewidth]{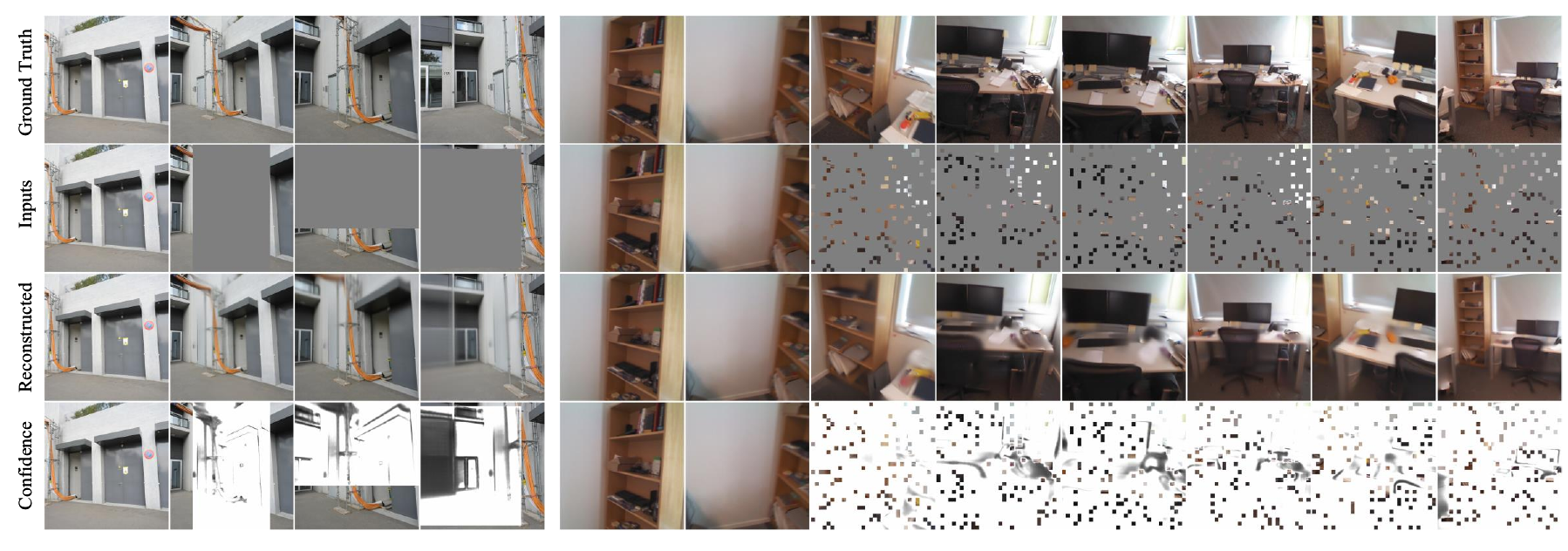}
    \caption{
    \textbf{Visualizations of the reconstructions from masked images.} 
    Samples are taken from ETH3D\cite{eth3d} and 7Scenes\cite{7scenes}. 
    % Left: Using a reference view, the model successfully reconstructs co-visible regions while assigning low confidence to areas that cannot be seen in the reference. 
    % Right: The model can also reconstruct a coherent scene by aggregating sparse, fragmented cues from all views.
    Left: The model successfully reconstructs co-visible regions. Meanwhile, for areas that are not co-visible (e.g., newly exposed surfaces due to camera motion), the model produces a blurry reconstruction and assigns a low confidence score. 
    Right: In a more challenging scenario where the reference views provide little information, the model can aggregate information from sparse cues to reconstruct.
    }
    \label{fig:recovered}
    \vspace{-0.3cm}
\end{figure*}
    
    % Brighter color indicates higher confidence. In the left examples with rectangular masks, non-co-visible regions are reconstructed with blurry details and low confidence. In the right examples with random masks, even when the reference views provide little direct information, the randomly distributed visible patches still offer structural cues that enable successful reconstruction.
In this section, we present \themethod, a self-supervised pre-training framework designed to learn 3D-aware visual representations. The core principle of \themethod is to train a model to solve a challenging multi-view completion task. By learning to reconstruct heavily masked content in one view using information from other views, the model is implicitly forced to find geometric correspondences and develop view-invariant features. We detail the design of pretext task, model architecture, and training objective below.

\subsection{Pretext Task}

At the core of our method is a pretext task designed to solve a challenging \textit{multi-view completion task}.
Formally, let $\mathcal{X} = \{x_1, x_2, \ldots, x_V\}$ be a collection of $V$ images of the same scene captured from different viewpoints. 
Each image $x_i \in \mathcal{X}$ is first divided into a sequence of $N$ non-overlapping patches, denoted as $\{p_{i,j}\}_{j=1}^N$. 
For each view $i$, we randomly mask a large portion of its patches. Let $\mathcal{M}$ be the set of indices $(i,j)$ corresponding to all masked patches across all views, with $r \in [0, 1]$ being a masking ratio hyperparameter.
The objective of our pretext task is to reconstruct the original pixel values of all masked patches $\{p_{i,j} | (i,j) \in \mathcal{M}\}$ by observing the remaining visible patches from \textit{all} views, as shown in \cref{fig:method}.
% We typically use $r = 0.9$, i.e., $90\%$ of patches will be masked.

This formulation marks a fundamental departure from conventional single-view Masked Image Modeling~(MIM). While single-view MIM encourages the learning of semantic and textural regularities within a single image (e.g., reconstructing a cat's ear from its face), it primarily relies on contextual reasoning.
In our setting, however, we make reconstruction from single-view context nearly impossible by masking large, contiguous regions.
This design compels the model to shift from contextual reasoning to geometric reasoning. 
To reconstruct a heavily masked patch in view B, it must identify and utilize the corresponding patch from another view A. In essence, the model is implicitly forced to answer the question: \textit{What patch in view A corresponds to this masked patch in view B?} 
This shift is the cornerstone of \themethod's ability to learn 3D-aware representations without requiring any explicit 3D supervision.

\subsection{Model Design}

\paragraph{Architecture Design.}
To jointly process patch tokens from multiple views, we adopt stacked Alternating Attention~(AA) blocks following the design of VGGT~\cite{wang2025vggt}. Each block alternates between frame-wise and global attention, enabling hierarchical aggregation of intra- and inter-view information. In particular, the global attention operates across views through cross-view query–key interactions, effectively associating spatially corresponding regions between different viewpoints. This mechanism implicitly establishes correspondences, thereby encouraging the emergence of multi-view consistent feature representations. Unlike VGGT~\cite{wang2025vggt}, which distinguishes between primary and secondary views, \themethod treats all views equally and ensures permutation equivariance. We further incorporate Rotary Positional Embeddings~(RoPE)~\cite{su2024roformer} to enhance resolution adaptability. A lightweight linear head is used to decode pixel values and confidence maps during pre-training and is discarded afterwards.

\begin{comment}
\paragraph{Lightweight Decoder. \lsd{Consider deleting this paragraph.}}
A key principle of our design is that the encoder should bear the full responsibility of learning powerful, semantically and geometrically rich features. To this end, we employ a lightweight and simple linear decoder, which takes the encoded patch tokens as input and projects them back to the pixel space. This minimal design prevents the decoder from becoming a powerful "crutch" that could hallucinate content without true understanding. It forces the encoder to produce features that are already well-structured for the reconstruction task, thereby ensuring the quality of the learned representations.
\end{comment}

\paragraph{Masking Strategy.}

The design of our masking strategy is driven by a primary objective: to eliminate \textit{shortcuts}. We define a shortcut as any mechanism by which the model can reconstruct masked content using only information from a single view, thereby bypassing the need to learn multi-view geometric correspondences. 
We introduce three key principles to achieve this.
First, we employ a high masking ratio. A low ratio would create an obvious shortcut: the model could simply rely on the rich local context within a single frame for reconstruction. Our masking ratio is considerably higher than the 75\% used in MAE\cite{he21masked}.
Second, besides random masks with high ratio, we apply spatially concentrated masks (e.g., large contiguous blocks) instead of random per-patch masking only.
This design prevents the model from leveraging sparse structure cues that can persist in a single, randomly masked view. By masking large, contiguous regions, we make single-view reconstruction nearly impossible, forcing the model to seek information from other views. Specifically, we apply rectangular and elliptical region masks as shown in \cref{fig:method}.
Thirdly, a subset of views remain unmasked to serve as reference. 
The ablation study in ~\cref{tab:ablation_mask} validates these three principles.
% , we find that the best performance is achieved when a high 0.90 masking ratio is combined with our mask strategy and a single reference view.

\subsection{Training}
% \paragraph{Objective Function.}
\paragraph{Objective Function.}
The training objective is to minimize the reconstruction error in the pixel space for the masked patches. Since certain masked regions may not contain adequate contextual information for reliable reconstruction, we adopt a confidence-aware $\mathcal{L}_2$ loss to adaptively weight the reconstruction objective. In addition to stabilizing training, the learned confidence maps implicitly encode cross-view correspondence cues, which further benefit multi-view understanding. 
For each patch in the masked set $\mathcal{M}$, the model outputs both the reconstructed pixel values $\hat{p}_{i,j}$ and a corresponding confidence score $c_{i,j}$, which is normalized to $[0,1]$ by a sigmoid function.
The confidence-aware $\mathcal{L}_2$ loss is formally defined as:
\begin{equation}
    \label{eq:confloss}
    \mathcal{L} = \frac{1}{{|\mathcal{M}|}} \sum_{(i,j) \in \mathcal{M}} \left[(c_{i,j}+\epsilon)\|\hat{p}_{i,j} - p_{i,j} \|_2^2 - \lambda \log c_{i,j}\right],
\end{equation}
where $\epsilon$ and $\lambda$ are set to 0.1 in all experiments.

\begin{comment}
{This simple pixel-level objective, when coupled with our geometrically challenging pretext task, provides a powerful supervisory signal. 
It effectively guides the model to learn high-level, 3D-aware feature representations that are robust to viewpoint changes, without requiring any explicit 3D ground truth data.}
\end{comment}

\begin{figure*}[!htbp]
    \centering
    \includegraphics[width=\linewidth]{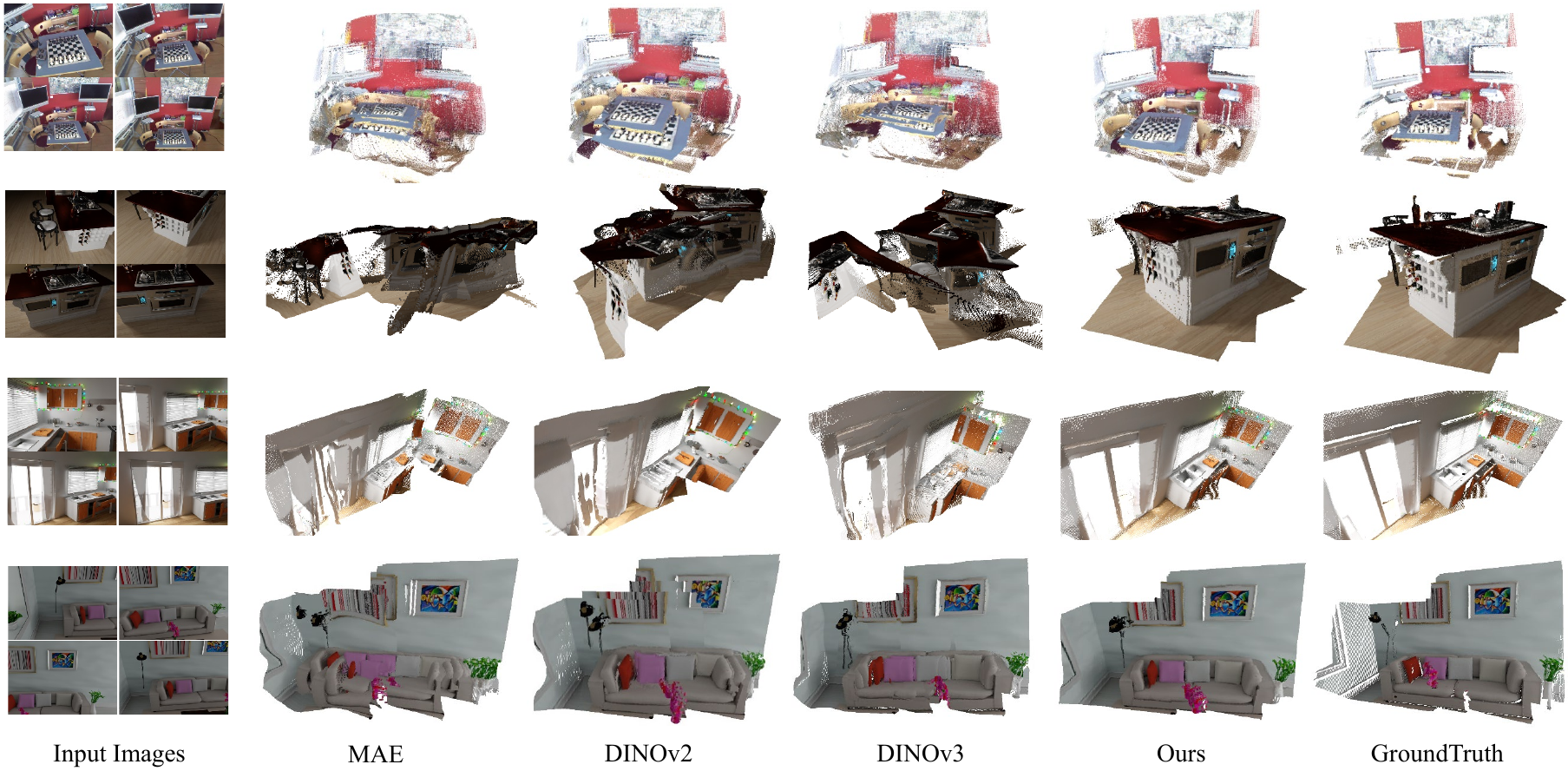}
    \caption{
    \textbf{Qualitative comparison of predicted pointmaps for 7Scenes\cite{7scenes} and NRGBD\cite{nrgbd_dataset_cvpr22}.}
    We compare the 3D reconstruction results of our method against baselines using different visual backbones.
    Our method consistently produces reconstructions that more complete and more faithful to the ground truth geometry across all scenes.
    }
      % Compared to other visual backbones, M$^2$AE produces cleaner, more accurate and more complete reconstructions with fewer artifacts.}
    \label{fig:pts}
% \vspace{-0.3cm}
\end{figure*}
\paragraph{Training Data.}
We pre-train \themethod on a large-scale and diverse collection of multi-view datasets, comprising a mixture of: Co3Dv2~\cite{reizenstein21common}, BlendedMVG~\cite{yao2020blendedmvs}, ARkitScenes~\cite{dehghan2021arkitscenes}, DL3DV~\cite{ling2024dl3dv}, MegaDepth~\cite{li2018megadepth}, ScanNet++~\cite{yeshwanthliu2023scannetpp}, HyperSim~\cite{hypersim}, Waymo~\cite{waymo} and RealEstate10K\cite{zhou2018stereo}.
This curated collection spans a wide range of domains—from indoor and outdoor scenes to synthetic and real-world captures—ensuring the model learns generalizable features. 
In total, our pre-training dataset is comparable in scale and diversity to that used by recent feed-forward 3D Reconstruction models such as VGGT\cite{wang2025vggt}.
% The 3D annotations for these datasets are derived from multiple sources, such as direct sensor capture, synthetic engines, or SfM techniques.

\paragraph{Implementation Details.}
% In pre-training, we employ an AdamW\cite{kingma14adam:} optimizer with a cosine learning rate scheduler, and train for 400 epochs.
% % In self-supervised pre-training, we employ an AdamW optimizer with a cosine learning rate scheduler, and train for 100 epochs. 
% The training hyper-parameters are: the batch size as 2048, base learning rate as 8e-4, weight decay as 0.05, $\beta_1$ = 0.9, $\beta_2$ = 0.999, warm-up for 10 epochs. A light data augmentation strategy is used: random resize cropping with scale range of [0.67, 1] and a aspect ratio range of [3/4, 4/3], followed by a random flipping and a color normalization steps.
% hybrid resolution, various num views, nearby sampling, training details
During pre-training, the model is trained with mixed image resolutions and aspect ratios, including 224, 384, and 512 pixels. 
The number of input views is randomly varied between 2 and 8 for each training sample to enhance robustness across different multi-view configurations. 
We adopt the AdamW\cite{kingma14adam:} optimizer with a learning rate of $2\times 10^{-4}$, following a cosine decay schedule and a 2-epoch warm-up. 
The pre-training runs for 400 epochs with 200K randomly sampled image groups per epoch. 
For each image group, we select adjacent views based on image IDs to improve sample effectiveness.
Standard data augmentations, including random cropping and flipping, are applied to improve generalization.
We train two model sizes, \themethod-B and \themethod-L, with parameter counts equal to the standard ViT-Base and ViT-Large, respectively.
The training runs on 8 A100 GPUs about two weeks for \themethod-L and one week for \themethod-B.
We provide visualization of two challenging scenes unseen during pre-training, which shows strong evidence that the model has learned to solve the multi-view completion task.

% \paragraph{Qualitative Visualization.}  visualizes .
% In the left example, where one view is kept as an unmasked reference, our model successfully reconstructs heavily masked regions in the other views. Crucially, for areas that are not co-visible in the reference view (e.g., newly exposed surfaces due to camera motion), the model produces a blurry reconstruction and, correspondingly, assigns a low confidence score. 
% The right example  presents a more challenging scenario where the reference views provide little direct information. In this setting, the model can aggregate geometric information from sparse visual cues to reconstruct.
% This behavior provides strong evidence that the model learns to solve the task by finding geometric correspondences across views.
\section{Experiments}
\begin{table*}[t!]
\setlength{\tabcolsep}{3pt}
\centering
\small
\resizebox{\linewidth}{!}{
\begin{tabular}{@{} l l *{3}{c} *{3}{c} *{3}{c} *{4}{c} @{}}
\toprule
\multirow{2}{*}{Dataset} & \multirow{2}{*}{Method} & 
\multicolumn{3}{c}{$R@k\uparrow$} & 
\multicolumn{3}{c}{$T@k\uparrow$} & 
\multicolumn{3}{c}{$AUC@k\uparrow$} & 
\multicolumn{4}{c}{Pointmap Estimation$\downarrow$} \\
\cmidrule(lr){3-5} \cmidrule(lr){6-8} \cmidrule(lr){9-11} \cmidrule(l){12-15}
 & & @5 & @15 & @30 & @5 & @15 & @30 & @5 & @15 & @30 & Acc & Comp & Overall & $||\mathcal{L}_1||$ \\
\midrule
\multirow{6}{*}{7Scenes} 
 & MAE-L\cite{he21masked} & 22.571 & 50.071 & 62.786 & 0.857 & 6.357 & 20.000 & 0.029 & 1.081 & 4.536 & 0.055 & 0.107 & 0.081 & 0.129 \\
 & DINOv2-L\cite{oquab24dinov2:} & 49.143 & 81.071 & 91.000 & 0.714 & 7.929 & 23.357 & 0.186 & 2.181 & 8.514 & \underline{0.039} & 0.042 & 0.041 & 0.074 \\
 & DINOv3-L\cite{simeoni2025dinov3} & 38.929 & 65.500 & 72.214 & 1.929 & 10.357 & 25.786 & 0.314 & 3.729 & 9.826 & 0.050 & 0.092 & 0.071 & 0.115 \\
 & \themethod-B & \underline{74.214} & \underline{97.357} & \underline{98.786} & \underline{5.857} & \underline{32.857} & \underline{60.071} & \underline{1.414} & \underline{13.810} & \underline{31.231} & \underline{0.039} & \underline{0.041} & \underline{0.040} & \underline{0.051} \\
 & \themethod-L & \textbf{90.143} & \textbf{99.214} & \textbf{100.000} & \textbf{16.286} & \textbf{52.143} & \textbf{78.571} & \textbf{6.100} & \textbf{26.929} & \textbf{47.345} & \textbf{0.025} & \textbf{0.028} & \textbf{0.027} & \textbf{0.035} \\
 \cmidrule{2-15}
 & $\pi^3$(ref.)\cite{wang2025pi3} & 94.571 & 100.000 & 100.000 & 37.643 & 74.214 & 86.857 & 16.200 & 46.029 & 64.069 & 0.014 & 0.016 & 0.015 & 0.022 \\
\midrule
\multirow{6}{*}{NRGBD}
 & MAE-L\cite{he21masked} & 40.286 & 62.143 & 75.786 & 2.571 & 15.071 & 36.643 & 0.400 & 4.276 & 12.781 & 0.073 & 0.116 & 0.095 & 0.129 \\
 & DINOv2-L\cite{oquab24dinov2:} & 60.500 & 82.143 & 87.714 & 5.071 & 30.500 & 56.643 & 1.057 & 10.238 & 25.986 & 0.060 & \underline{0.089} & 0.075 & 0.096 \\
 & DINOv3-L\cite{simeoni2025dinov3} & 45.714 & 73.000 & 77.857 & 4.500 & 18.357 & 41.143 & 0.857 & 6.229 & 16.195 & 0.064 & 0.101 & 0.083 & 0.114 \\
 & \themethod-B & \underline{70.643} & \underline{88.357} & \underline{89.643} & \underline{13.571} & \underline{53.571} & \underline{77.000} & \underline{3.686} & \underline{25.038} & \underline{45.433} & \underline{0.058} & 0.090 & \underline{0.074} & \underline{0.090} \\
 & \themethod-L & \textbf{92.714} & \textbf{94.143} & \textbf{94.214} & \textbf{41.571} & \textbf{80.071} & \textbf{89.857} & \textbf{19.457} & \textbf{50.957} & \textbf{67.452} & \textbf{0.046} & \textbf{0.077} & \textbf{0.061} & \textbf{0.067} \\
\cmidrule(lr){2-15}
 & $\pi^3$(ref.)\cite{wang2025pi3} & 94.714 & 95.071 & 95.143 & 79.286 & 92.071 & 96.000 & 57.800 & 77.495 & 84.843 & 0.031 & 0.060 & 0.046 & 0.044 \\
\bottomrule
\end{tabular}
}
\caption{
\textbf{Quantitative results on 7Scenes\cite{7scenes} and NRGBD\cite{nrgbd_dataset_cvpr22} datasets for 3D reconstruction}. These two dataset are not trained in pre-training and downstream finetuning. Best results are in bold.
}
\label{tab:3Dbenchmark}
\end{table*}

\begin{comment}
In this section, we conduct a series of extensive experiments to validate the effectiveness of \themethod pre-training. 
Our evaluation is structured to answer three key questions: (1) Does \themethod, when used as a general-purpose feature backbone, lead to superior performance on challenging downstream 3D geometric tasks compared to state-of-the-art alternatives? (2) Can we find direct evidence that \themethod learns features with better geometric consistency and correspondence quality, thus verifying our core hypothesis? (3) How do our key design choices, such as the masking strategy and model architecture, contribute to the final performance? We first evaluate on downstream tasks, then present a direct analysis of correspondence quality, and conclude with in-depth ablation studies. \lsd{too many and too long questions? Consider changing them to chenshuju.}
\end{comment}
We evaluate \themethod pre-training through a threefold analysis.
% First, we assess its performance on downstream 3D geometric tasks against state-of-the-art methods. 
% Second, we directly analyze the geometric consistency of its learned features. 
First, we analyze the geometric consistency of its learned features. 
Second, we assess its performance as the backbone for downstream 3D geometric tasks against state-of-the-art methods.
Finally, we conduct ablation studies to investigate the impact of our key design choices.

\subsection{Zero-Shot Correspondence Quality}
\begin{table}[t!]
\centering
\small
\resizebox{\columnwidth}{!}{
\begin{tabular}{@{} l| ccc |ccc @{}}
\toprule
\multirow{2}{*}{Method} & \multicolumn{3}{c|}{Pixel Space~(px)} & \multicolumn{3}{c}{3D Space~(cm)} \\
\cmidrule{2-4} \cmidrule{5-7}
 & ATE $\downarrow$ & @5 $\uparrow$ & @10 $\uparrow$ & ATE $\downarrow$ & @2 $\uparrow$ & @5 $\uparrow$ \\
\midrule
CLIP\cite{radford21learning} & 189.22 & 12.97 & 13.35 & 13.86 & 23.84 & 41.10 \\
SigLIP\cite{siglip} & 111.66 & 13.66 & 15.71 & 7.79 & 43.46 & 68.41 \\
\cmidrule{1-7}
SAM\cite{sam} & 69.54 & 15.68 & 22.28 & 4.56 & 62.58 & 81.66 \\
\cmidrule{1-7}
DIFT\cite{dift} & 60.44 & 16.97 & 26.46 & 4.93 & 64.36 & 81.21 \\
MAE\cite{he21masked} & 81.24 & 14.63 & 19.11 & 5.25 & 54.73 & 78.19 \\
CroCov2\cite{weinzaepfel2023crocov2} & 56.79 & \underline{19.77} & \underline{32.58} & 4.42 & 72.15 & 84.66 \\
DINOv2\cite{oquab24dinov2:} & 53.31 & 16.94 & 26.29 & 4.24 & 69.97 & 84.23 \\
DINOv3\cite{simeoni2025dinov3} & \underline{41.15} & 17.23 & 27.95 & 3.76 & 74.89 & 87.11 \\
\themethod-B & 42.54 & {19.48} & {31.95} & \underline{2.93} & \underline{75.90} & \underline{89.17} \\
\themethod-L & \textbf{29.42} & \textbf{22.29} & \textbf{39.21} & \textbf{2.38} & \textbf{82.74} & \textbf{91.87} \\
\bottomrule
\end{tabular}
}
\caption{
\textbf{Quantitative results on the NAVI\cite{jampani2023navi} dataset for zero-shot correspondence.} 
% We compare our \themethod models against various foundational models. 
% \TODO{compare to zeroco}
}
\label{tab:navi_track_results}
% \vspace{-0.3cm}
\end{table}
The core idea of \themethod is to learn geometrically consistent features. We validate this directly by evaluating the model's zero-shot correspondence quality, isolating the effect of our pre-training from any downstream fine-tuning.
% Multi-view consistency plays a decisive role in 3D perception. Here, we evaluate \themethod’s ability in this regard through correspondence-based metrics.
% Feature consistency across views is fundamental to 3D geometric understanding. 
% To test this, we evaluate the quality of its geometric correspondences in a zero-shot setting. 
% This directly measures the geometric fidelity of the learned features, isolating the encoder's performance from any downstream components.
% To directly assess the geometric consistency of our learned features, we evaluate their performance on a challenging zero-shot point correspondence task. 
% This setup requires correspondence a set of points across a sequence of multiple views without any fine-tuning.
% The cornerstone of 3D geometric understanding is the ability to maintain feature consistency across multiple views. Our central claim is that \themethod's pre-training objective—reconstructing masked regions by leveraging other views—directly forces the model to learn such view-invariant representations. Therefore, to directly test this hypothesis, we now evaluate the quality of geometric correspondences in a zero-shot manner. This provides a direct, fundamental measure of the geometric fidelity of the learned features, isolating the encoder's contribution from any downstream components.

\begin{table}[t!]
\centering
\small
\resizebox{\columnwidth}{!}{
% --- 在列定义中加入竖线 ---
\begin{tabular}{@{} l|ccc|ccc @{}}
\toprule
% --- 使用 multirow 来“强制”画竖线 ---
\multirow{2}{*}{Method} & \multicolumn{3}{c|}{Pixel Space~(px)} & \multicolumn{3}{c}{3D Space~(cm)} \\
\cmidrule(lr){2-4} \cmidrule(lr){5-7}
 & ATE~$\downarrow$ & @5~$\uparrow$ & @10~$\uparrow$ & ATE~$\downarrow$ & @2~$\uparrow$ & @5~$\uparrow$ \\
\midrule
CLIP\cite{radford21learning} & 203.18 & 20.83 & 21.04 & 102.45 & 22.47 & 22.95 \\
SigLIP\cite{siglip} & 142.96 & 21.04 & 21.78 & 81.44 & 23.04 & 24.88 \\
\cmidrule{1-7}
SAM\cite{sam} & 76.12 & 23.80 & 31.16 & 49.85 & 27.34 & 36.56 \\
\cmidrule{1-7}
MAE\cite{he21masked} & 104.94 & 22.07 & 25.18 & 62.70 & 24.01 & 28.51 \\
CroCov2\cite{weinzaepfel2023crocov2} & 25.92 & \textbf{34.20} & \textbf{55.93} & 29.84 & \textbf{38.16} & \textbf{58.02} \\
DINOv2\cite{oquab24dinov2:} & 72.11 & 23.42 & 29.22 & 49.09 & 27.13 & 34.29 \\
DINOv3\cite{simeoni2025dinov3} & 55.78 & 24.66 & 33.54 & 42.58 & 29.49 & 41.69 \\
\themethod-B  & 33.77 & 31.41 & 47.67 & 31.67 & 35.03 & 51.14 \\
\themethod-L & \textbf{24.68} & \underline{33.83} & \underline{52.96} & \textbf{27.82} & \underline{37.52} & \underline{56.60} \\
\bottomrule
\end{tabular}
}
\caption{
\textbf{Quantitative results on the ScanNet\cite{dai2017scannet} dataset for zero-shot correspondence.}
% We compare our \themethod models against various foundational models. 
% \TODO{compare to zeroco}
}
\label{tab:scannet_track_results}
% \vspace{-0.3cm}
\end{table}
\paragraph{Implementations.}
Our evaluation is performed on multi-view image sequences, each consisting of 8 frames. For each sequence, we sample a set of points in the first frame and track their corresponding 2D locations in all subsequent frames. The performance is then measured by the error between the predicted tracks and the ground-truth tracks.
A key difference in this evaluation lies in how correspondences are extracted from our native multi-view model versus the frame-wise baselines.
% To directly assess the geometric consistency of our learned features, we evaluate their performance on a challenging zero-shot point correspondence task. This setup, which requires correspondence points across multiple views without any fine-tuning, provides a pure measure of the encoder's geometric fidelity, isolating it from any downstream components.
% To directly assess the geometric consistency of our learned features, we evaluate their performance on a challenging zero-shot point correspondence task. This setup requires correspondence a set of points across a sequence of multiple views without any fine-tuning.
For \themethod, we perform a single forward pass over all views and extract attention maps from global attention layer, from which a dense correlation volume is regressed using a \texttt{soft-argmax} operation to extract correspondence.
% Inspired by recent findings~\cite{an2025zeroco}, we symmetrize the bidirectional attention between the first and subsequent views to create a robust correlation volume, from which a dense flow field is regressed using a \texttt{soft-argmax} operation to predict tracks.
For baseline frame-wise models, we extract dense features for each view. Correspondences are then established via nearest-neighbor matching in the feature space. 
% All methods are evaluated in a zero-shot manner, directly using their pretrained weights without any task-specific adaptation. 
% Further implementation details are provided in the Appendix.

% We assess consistency without additional training by directly computing dense correspondences from the feature maps extracted by each encoder. This zero-shot procedure follows keypoint-free pipelines and aligns with recent feature quality benchmarks. Specifically, for a pair of images, we extract dense patch-level features using the frozen encoder (ViT-L/14 resolution). 
% Correspondences are established via mutual nearest-neighbor matching in feature space, thresholded by cosine similarity (>0.7) and refined with subpixel accuracy via soft-argmax. This yields pixel- and 3D-level error distributions, allowing direct evaluation of geometric fidelity without probes.

\paragraph{Datasets and Metrics.}
Following the setup in~\cite{probe3d}, we evaluate on both indoor scenes from Paired ScanNet split\cite{dai2017scannet, sarlin20superglue:}\footnote{The ScanNet splits used for evaluation have no overlap with the ScanNet++ data used during our pre-training.} and objects from NAVI\cite{jampani2023navi}. 
We report performance using two primary metrics: (1) 2D pixel error, which includes the Average Trajectory Error (ATE) in pixels and accuracy at various pixel thresholds (Acc@k px); and (2) 3D metric error, where we unproject points into 3D space to compute ATE in centimeters and accuracy at various centimeter thresholds (Acc@k cm). 
We only consider points that are visible in the ground truth for a fair evaluation. 
% This comprehensive set of metrics allows us to rigorously assess both the 2D image-plane consistency and the 3D geometric fidelity of the learned representations.
% Further details on the datasets and metric computation are provided in the Appendix.

\begin{figure*}[htbp]
    \centering
    \includegraphics[width=\linewidth]{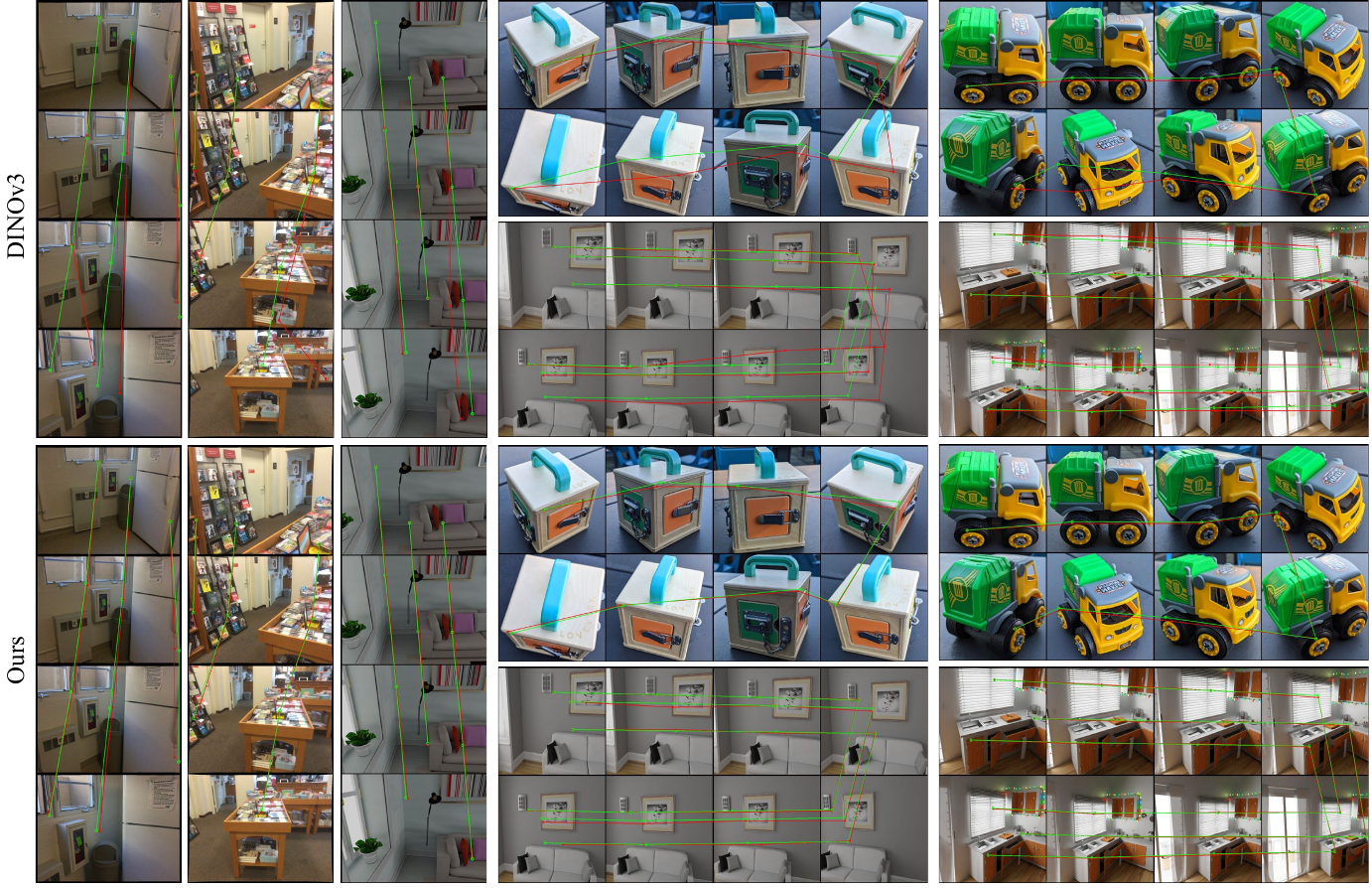}
    \caption{
    \textbf{Qualitative comparison of multi-view correspondence estimation.}
     Sequences are sampled from the NAVI\cite{jampani2023navi}, ScanNet\cite{dai2017scannet}, and NRGBD\cite{nrgbd_dataset_cvpr22}.
     We visualize the predicted tracks (red) and the ground-truth tracks (green).
     Points are only rendered in frames where the GT point is visible.
     The close alignment between the red and green paths highlights our method's ability to maintain accurate consistency.
    }
    \label{fig:correspondence}
\end{figure*}

\paragraph{Results.}
The quantitative results, presented in \cref{tab:navi_track_results,tab:scannet_track_results}, highlight the superior multi-view consistency of our \themethod models across both objects and scenes.\footnote{All baseline VFMs are in their Large variants, except for CroCov2 which uses a Large encoder and a Base decoder.}
On the object-centric NAVI\cite{jampani2023navi} dataset, \mbox{\themethod-L} achieves a remarkable 3D-space ATE of just 2.38 cm, a 36\% error reduction over the DINOv3.
Our model’s performance advantage is also pronounced on the ScanNet\cite{dai2017scannet} dataset, where the smaller \mbox{\themethod-B} model outperforms all frame-wise baselines.
% CroCo\cite{weinzaepfel2023crocov2} is predominantly trained on indoor-scene datasets, which biases the model toward stronger geometric reasoning in indoor environments (e.g., ScanNet\cite{dai2017scannet}). Consequently, its performance degrades on object-centric scenarios such as NAVI\cite{jampani2023navi}.
% The consistent performance gain across both datasets validates that our pre-training approach successfully instills a robust geometric consistency.
CroCo\cite{weinzaepfel2023crocov2} performs well on ScanNet\cite{dai2017scannet} but generalizes poorly to object-centric NAVI\cite{jampani2023navi} due to its indoor-focused training data. In contrast, our models show consistent gains across both datasets, demonstrating stronger geometric generalization.
Meanwhile, we present a qualitative comparison of predicted correspondences in \cref{fig:correspondence}. We visualize the predicted tracks in red and the ground-truth tracks in green. The close alignment between the red and green paths highlights \themethod's ability to maintain high-quality correspondence even in texture-less areas.

\subsection{Performance on 3D Reconstruction Tasks}
In this section, we evaluate \themethod's utility as a general-purpose backbone for 3D vision. We replace the standard encoder in a state-of-the-art 3D reconstruction framework with ours and measure the resulting performance gains.
% We first evaluate \themethod on multi-view 3D reconstruction tasks as it is a quintessential test for 3D-aware vision models. Success in this task is fundamentally dependent on the ability to find and aggregate geometric correspondences across views—the exact skill \themethod is pre-trained to acquire through its multi-view inpainting objective. Therefore, this task serves as a direct and practical measure of how well \themethod's learned 3D inductive biases transfer to a complex downstream application.
% As a cornerstone of computer vision, it demands precise integration of cross-view geometric cues—such as parallax and epipolar constraints—to infer scene structure, directly aligning with \themethod's self-supervised objective of learning view-invariant features through multi-view inpainting. 
% Unlike single-view tasks (e.g., classification), 3D reconstruction inherently requires aggregating inter-view correspondences, making it an ideal proxy to demonstrate the transferability of \themethod's implicit 3D inductive biases.
% In this section, we substitute \themethod as the encoder in feed-forward pipelines and study how ours models transfer to 3D reconstruction tasks, including depth, pointmaps, and camera poses estimation. 

\paragraph{Implementations.}
We evaluate \themethod\ by integrating it as the feature backbone of the state-of-the-art $\pi^3$~\cite{wang2025pi3} reconstruction pipeline, which offers permutation equivariance and faster convergence than VGGT~\cite{wang2025vggt}.
To better isolate the encoder’s contribution, we simplify 36-layer decoder of $\pi^3$ to a lightweight 4-layer version.
For all experiments, we replace the original DINOv2~\cite{oquab24dinov2:} backbone in this simplified framework with different frame-wise ViTs like MAE\cite{he21masked}, DINOv3~\cite{simeoni2025dinov3}, and \themethod series, while keeping all other components and hyperparameters identical.
This setup enables a fair comparison focused solely on the encoder’s impact on reconstruction quality.

\begin{comment}
However, the original $\pi^3$ architecture employs an extremely deep 36-layer decoder, \lsddel{which is computationally prohibitive for our academic research setting and makes}\lsd{making} it difficult to isolate the true contribution of the feature encoder.
To address this\lsddel{ and to create a more focused and efficient evaluation testbed}, we make a simplification by reducing the decoder depth from 36 layers to a much more lightweight 4-layer design.
\end{comment}

\paragraph{Training and Evaluation Protocol.} 
We train all downstream models on ARKitScenes~\cite{dehghan2021arkitscenes}, ScanNet++~\cite{yeshwanthliu2023scannetpp}, and BlendedMVS~\cite{yao2020blendedmvs}, forming a lighter setup than the full corpus used by $\pi^3$~\cite{wang2025pi3}.
We evaluate performance on the 7Scenes\cite{7scenes} and NeuralRGBD\cite{nrgbd_dataset_cvpr22} benchmarks. 
For each scene, we randomly sample 8 frames. 
The predicted point cloud is aligned to the ground truth using the Umeyama\cite{umeyama1991least} algorithm.
For pointmap reconstruction, we report the smallest L2-distance between prediction to ground truth as Accuracy, the smallest L2-distance from ground truth to prediction as Completeness and their average as Overall.
We also report the mean absolute error ($||\mathcal{L}_1||$).
Lower values in Acc., Comp., Overall, and $||\mathcal{L}_1||$ indicate better reconstruction quality.
For camera pose regression, we measure rotation and translation accuracy (R@K and T@K) at thresholds $K={5,15,30}$, together with the area under the accuracy curve (AUC@K).

\paragraph{Results.} 

The quantitative results are shown in~\cref{tab:3Dbenchmark}.
In both pointmap reconstruction and camera pose estimation, \themethod-powered models consistently outperform all other baselines.
For instance, on the 7Scenes\cite{7scenes}, \themethod substantially outperforms the widely-adopted DINOv2; it boosts the overall camera pose accuracy (AUC@30) from 8.514 to a remarkable 47.345, and reduces the pointmap $||\mathcal{L}_1||$ error from 0.074 to 0.035.
% The advantage of \themethod is even more pronounced in camera pose estimation, which requires globally consistent features across views. 
Notably, the smaller \mbox{\themethod-B} variant consistently surpasses the larger \mbox{DINOv2-L} and \mbox{DINOv3-L} across nearly all metrics.
In addition, while $\pi^3$~\cite{wang2025pi3} attains strong results with a 36-layer decoder, our \mbox{\themethod-L} with 4-layer decoder achieves highly competitive performance, particularly in camera pose estimation.
These results strongly suggest that \themethod's pre-training forces the model to learn features that are inherently geometrically consistent, enabling efficient transfer to reconstruct accurate 3D structures.
We present a qualitative comparison of estimated pointmaps in \cref{fig:pts}, which highlights how our method maintains geometric consistency across views, particularly in challenging areas with repeating or homogeneous textures, such as checkerboards, murals, and sofas.

\subsection{Ablation Studies}
\begin{table}[t!]
    \centering
    \footnotesize
    \setlength{\tabcolsep}{4pt} % Adjust column spacing for better readability
    \resizebox{\columnwidth}{!}{
    \begin{tabular}{lcccccc}
        \toprule
        & \multicolumn{3}{c}{{Camera Poses}} & \multicolumn{3}{c}{{Pointmap}} \\
        \cmidrule(lr){2-4} \cmidrule(lr){5-7}
        & {AUC@5} & {AUC@15} & {AUC@30} & {Acc} & {Comp} & {$\mathcal{L}_1$} \\
        \midrule
        w/o pre-train & 0.021 & 0.598 & 3.760 & 0.0718 & 0.1123 & 0.1484 \\
        w pre-train  & 2.671 & 18.305 & 37.413 & 0.0461 & 0.0543 & 0.0639 \\
        \bottomrule
    \end{tabular}}
    \caption{\textbf{Ablation on the effects of architecture and pre-training.}
    Metrics are averaged over the 7Scenes~\cite{7scenes} and NRGBD~\cite{nrgbd_dataset_cvpr22}.
     \themethod (w/o pre-train) variant reflects the architectural changes without \mbox{pre-training} benefits, while \themethod (w \mbox{pre-train}) incorporates the full \mbox{pre-training} process.
    }
    \label{tab:pretraining_or_arch}
    \vspace{-0.25cm}
\end{table}
\begin{table}[t!]
\footnotesize
\centering
\resizebox{\columnwidth}{!}{
\begin{tabular}{ccccc}
\toprule
Input Views & 3D@1cm $\uparrow$ & 3D@5cm $\uparrow$ & 2D@5px $\uparrow$ & 2D@25px $\uparrow$ \\
\midrule
    2 & 40.59 & 82.91 & 4.43 & 33.74 \\
    3 & 42.89 & 83.83 & 4.74 & 35.50 \\
    4 & 44.45 & 84.58 & 5.10 & 36.80 \\
    5 & 46.13 & 84.57 & 5.28 & 38.38 \\
    6 & 46.75 & \textbf{84.82} & \textbf{5.45} & 38.91 \\
    7 & \underline{46.97} & \underline{84.77} & 5.41 & \underline{39.26} \\
    8 & \textbf{47.36} & 84.74 & \underline{5.43} & \textbf{39.52} \\
\bottomrule
\end{tabular}
}
\caption{\textbf{Ablation on the number of context views for correspondence quality on NAVI\cite{jampani2023navi}}. We evaluate the correspondence recall between a fixed pair while varying the context number. 
% Increasing N consistently improves performance, especially for high-precision metrics like 3D@1cm and 2D@25px.
}
\label{tab:ablation_on_mv_navi}
\vspace{-0.25cm}
\end{table}
% \begin{table}[t!]
% \centering
% \footnotesize
% \resizebox{\columnwidth}{!}{
% \begin{tabular}{lcccc}
% \toprule
% \multicolumn{1}{c}{} & \multicolumn{2}{c}{Pixel Space Recall $\uparrow$} & \multicolumn{2}{c}{3D Space Recall $\uparrow$} \\
% \cmidrule(lr){2-3} \cmidrule(lr){4-5}
% Mask Strategy & 2D@10px & 2D@50px & 3D@2cm & 3D@5cm \\
% \midrule
% Random Mask (k=2) & 61.33 & 91.68 & 36.58 & 55.76 \\
% \midrule
% Various Mask (k=0) & 63.44 & 93.47 & 37.11 & 57.83 \\
% Various Mask (k=1) & \textbf{65.26} & \textbf{94.04} & \textbf{37.98} & \textbf{59.41} \\
% Various Mask (k=2) & 63.79 & 93.54 & 37.57 & 58.03 \\
% Various Mask (k=4) & 63.45 & 92.52 & 37.49 & 57.79 \\
% \bottomrule
% \end{tabular}
% }
% \caption{\textbf{Ablation on masking strategies on NRGBD\cite{nrgbd_dataset_cvpr22}}. We compare a baseline "Random Mask" strategy against our proposed "Various Mask" with a varying number of unmasked reference views ($k$).
% % The "Various Mask" strategy with a single reference view ($k=1$) achieves the best performance across all representative recall metrics, indicating an optimal balance between providing a stable reference and forcing the model to infer cross-view geometry.
% }
% \label{tab:mask_strategy_ablation}
% \vspace{-0.3cm}
% \end{table}
\begin{table}[t!] % 使用 table* 环境使其横跨双栏，以获得更多空间
\centering
\setlength{\tabcolsep}{2pt}
\resizebox{\columnwidth}{!}{
\begin{tabular}{@{}lcccc@{}}
\toprule
{Configuration} & {ATE 2D (px) ↓} & {Acc@10px ↑} & {ATE 3D (cm) ↓} & {Acc@2cm ↑} \\
\midrule
\multicolumn{5}{l}{\textit{Ablation on Masking Ratio (Mask Type: Random, Ref. Views: 2)}} \\
\quad 0.75 & 22.82 & 59.50 & 16.49 & 36.32 \\
\quad 0.80 & 17.94 & 61.21 & 13.97 & 36.85 \\
\quad 0.85 & 17.28 & 61.17 & 13.68 & 36.67 \\
\quad 0.90 & \textbf{16.60} & \textbf{61.33} & \textbf{12.89} & \textbf{36.58} \\
\quad 0.95 & 18.88 & 55.63 & 14.43 & 35.14 \\
\midrule
% --- 第2部分：对掩码类型的消融 ---
\multicolumn{5}{l}{\textit{Ablation on Mask Type, Ref. Views: 2)}} \\
\quad Random (Ratio: 0.90) & 16.60 & 61.33 & 12.89 & 36.58 \\
\quad Ours & \textbf{14.37} & \textbf{63.79} & \textbf{11.83} & \textbf{37.57} \\
\midrule
% --- 第3部分：对参考视图数量的消融 ---
\multicolumn{5}{l}{\textit{Ablation on Number of Reference Views (Mask Type: Ours)}} \\
\quad 0 (No reference) & 14.82 & 63.44 & 12.55 & 37.11 \\
\quad 1 & \textbf{13.81} & \textbf{65.26} & \textbf{11.27} & \textbf{37.98} \\
\quad 2 & 14.37 & 63.79 & 11.83 & 37.57 \\
\quad 4 & 15.54 & 63.45 & 12.68 & 37.49 \\
\bottomrule
\end{tabular}
}
\caption{
\textbf{Ablation on the masking ratio, strategy, and number of reference frames on NRGBD\cite{nrgbd_dataset_cvpr22}.}
The results show that our mask strategy with one single reference view yield the best performance.
}
\label{tab:ablation_mask}
\vspace{-0.3cm}
\end{table}

\paragraph{Pre-training or Architecture.}
To disentangle the benefits of our architectural design from those of our multi-view pre-training strategy, we conduct an ablation study comparing full, pre-trained \themethod model against an identical architecture with randomly initialized weights.
The results in \cref{tab:pretraining_or_arch}, reveal that pre-training is the main factor for performance gain and the architecture-only variant contributes negligibly. 
This demonstrates that while our architecture provides a capable foundation for multi-view processing, its potential is only fully realized through our self-supervised pre-training. 
% The performance gains are not simply architectural but are fundamentally driven by \lsddel{the powerful} geometric priors learned during the pre-training phase.
\begin{figure}[t!]
    \centering % 将图内的所有内容居中
    % 3. 创建第一个子图
    \begin{subfigure}[b]{0.495\columnwidth}
        \includegraphics[width=\linewidth]{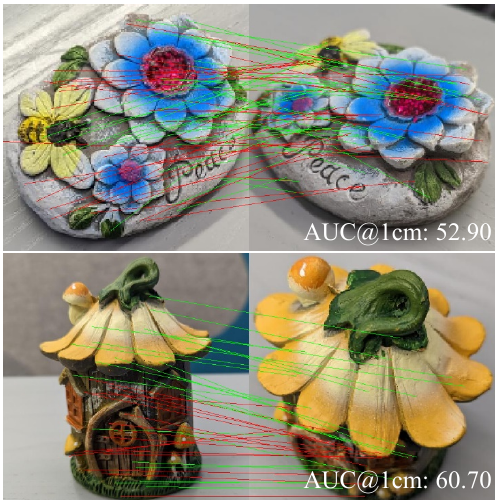}
        \subcaption{2 Input Views}
        \label{fig:image-a}
    \end{subfigure}
    % \hfill % 在两个子图之间添加一个弹性水平空间，将它们推向两边
    % 4. 创建第二个子图
    \begin{subfigure}[b]{0.495\columnwidth}
        \includegraphics[width=\linewidth]{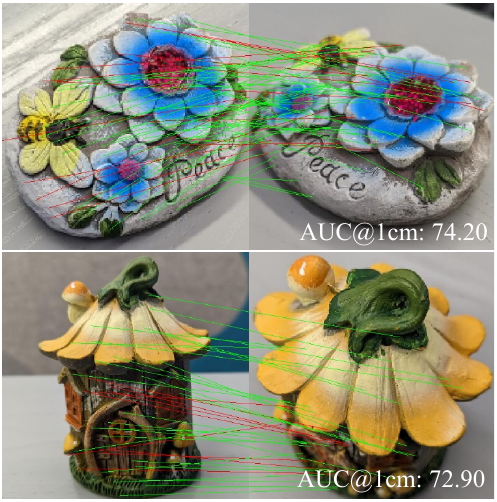}
        \subcaption{8 Input Views}
        \label{fig:image-b}
    \end{subfigure}
    % 5. 添加整个图的总标题和标签
    \caption{
        \textbf{Impact of multi-view context on correspondence quality from NAVI dataset~\cite{jampani2023navi}.} 
         We measure matching accuracy on a fixed image pair given (a) only the input pair (2 views) and (b) the pair plus six context views (8 views). Correct matches are shown in green and incorrect matches in red. The additional context yields performance gains, as evidenced by AUC@1cm score.
    }
    \label{fig:ablation_context_impact}
    \vspace{-0.5cm}
\end{figure}
\paragraph{Effect of Multi-view Context.}
% A core hypothesis of our work is that \themethod learns to perform genuine multi-view geometric reasoning. 
To verify that \themethod leverages multi-view context for reasoning, we conduct a controlled experiment in \cref{tab:ablation_on_mv_navi}. We measure correspondence accuracy on a fixed pair of images while varying the number of additional context views provided to the model. The results show a consistent trend: accuracy on the fixed pair improves as more context views are added. Notably, this performance gain is most pronounced in high-precision metrics like 3D@1cm. This experiment provides strong evidence that \themethod effectively uses information from additional views to refine its correspondence predictions.
We present a qualitative comparison in \cref{fig:ablation_context_impact}.
% To quantitatively validate this capability, we designed a controlled experiment to test whether providing additional, context-only views improves the correspondence accuracy between a fixed pair of images.
% The task is to establish dense correspondences between the first two views within an input image cluster. We vary the total number of views ($N$) provided as input to the model, from $N=2$ to $N=8$. The additional $N-2$ views serve purely as geometric context; the evaluation metrics are calculated exclusively on the first two views.

\paragraph{Masking Strategy.} 
We conduct an ablation study on the NRGBD\cite{nrgbd_dataset_cvpr22} to determine the optimal masking strategy.
We investigate three key aspects: masking ratio, mask type, and the number of reference views, as shown in \cref{tab:ablation_mask}. 
Our mask strategy randomly applies either a rectangular/elliptical mask (75\% ratio) or a random per-patch mask (90\% ratio) to each training sample.
The results show that using our proposed mask strategy with one reference view as final configuration yields the best overall performance.

\section{Conclusion}
This paper presents \themethod, a multi-view masked image modeling framework for 3D vision pre-training without relying on depth, camera pose, or any other annotations.
Our approach trains the model to reconstruct heavily masked parts of an image using information from other viewpoints and achieves higher geometric correspondences than existing visual backbones.
Furthermore, using \themethod as a feature extractor enhances the performance of downstream 3D tasks like 3D reconstruction.
This study demonstrates the potential of self-supervised learning for 3D vision. 
We believe 3D pre-training is promising and hope our work inspires progress in the computer vision community.

\appendix
% \clearpage
% \setcounter{page}{1}
% \maketitlesupplementary
\appendix
\newpage
\section*{Appendix}

\section{Experiments}
\subsection{Zero-Shot Correspondence Evaluation}
\begin{figure*}[t!]
    \centering
    
    \begin{minipage}[c]{0.03\linewidth}
        \centering
        \rotatebox{90}{DINOv3} 
    \end{minipage}
    \hfill
    \begin{minipage}[c]{0.96\linewidth}\includegraphics[width=\linewidth]{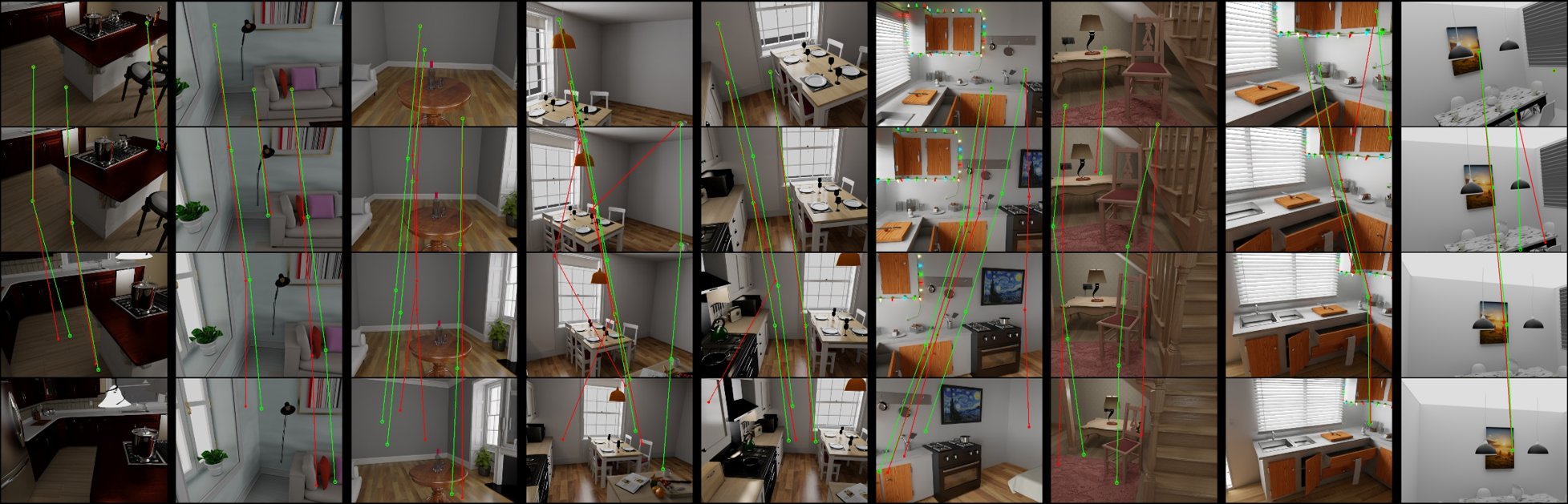}
    \end{minipage}
    
    \vspace{1mm}
    
    \begin{minipage}[c]{0.03\linewidth}
        \centering
        \rotatebox{90}{Ours} 
    \end{minipage}
    \hfill
    \begin{minipage}[c]{0.96\linewidth}\includegraphics[width=\linewidth]{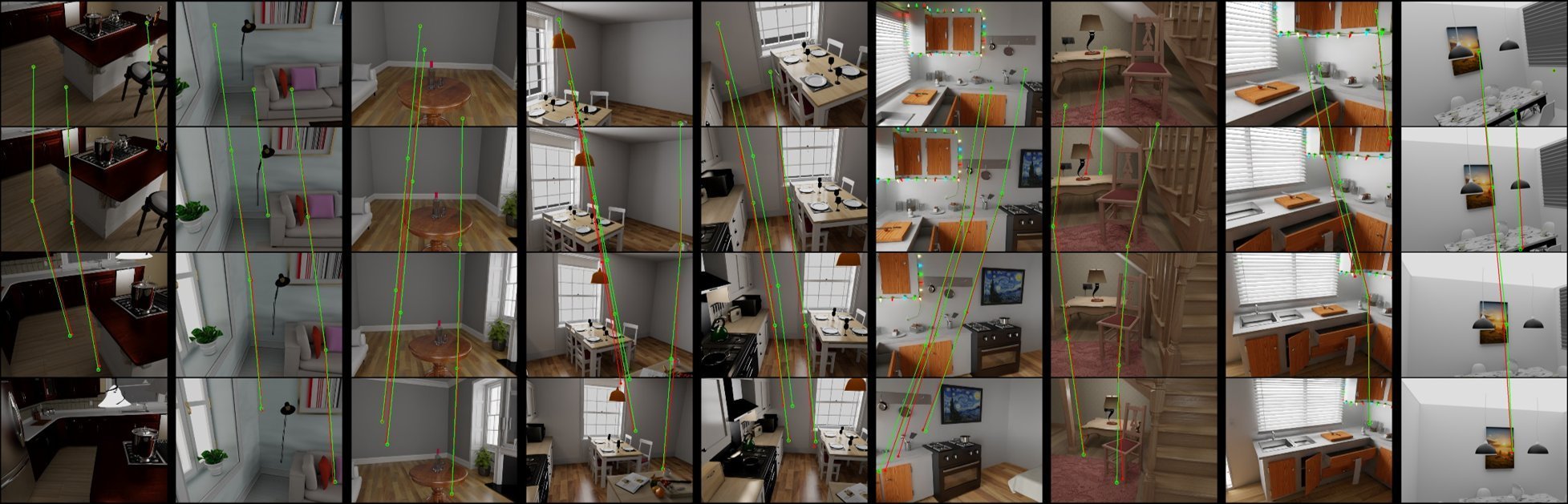}
    \end{minipage}

    \vspace{3mm}
    
    \begin{minipage}[c]{0.03\linewidth}
        \centering
        \rotatebox{90}{DINOv3} 
    \end{minipage}
    \hfill
    \begin{minipage}[c]{0.96\linewidth}\includegraphics[width=\linewidth]{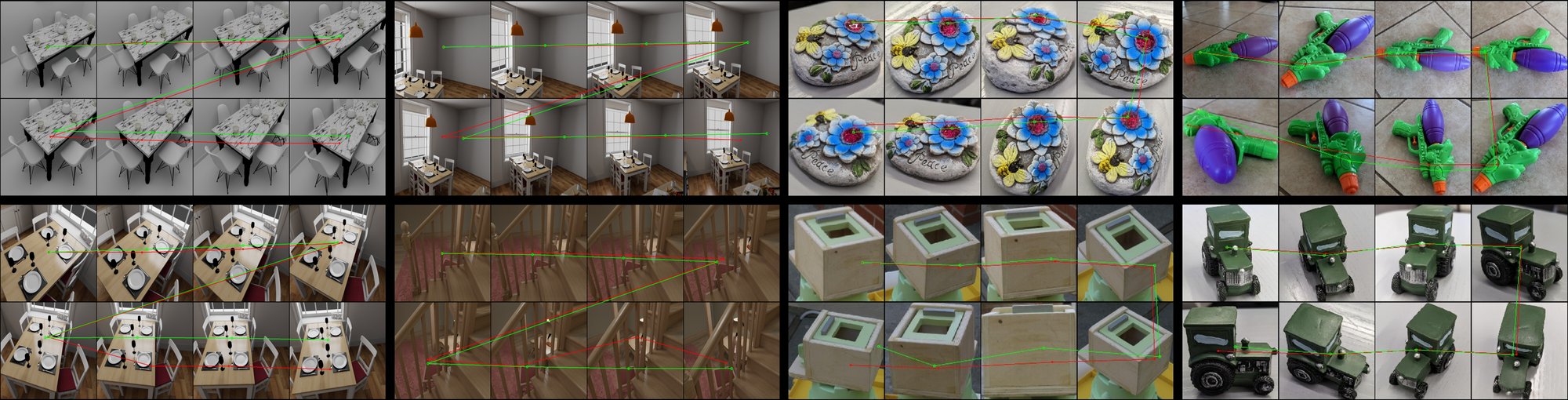}
    \end{minipage}
    
    \vspace{1mm}
    
    \begin{minipage}[c]{0.02\linewidth}
        \centering
        \rotatebox{90}{Ours} 
    \end{minipage}
    \hfill
    \begin{minipage}[c]{0.955\linewidth}\includegraphics[width=\linewidth]{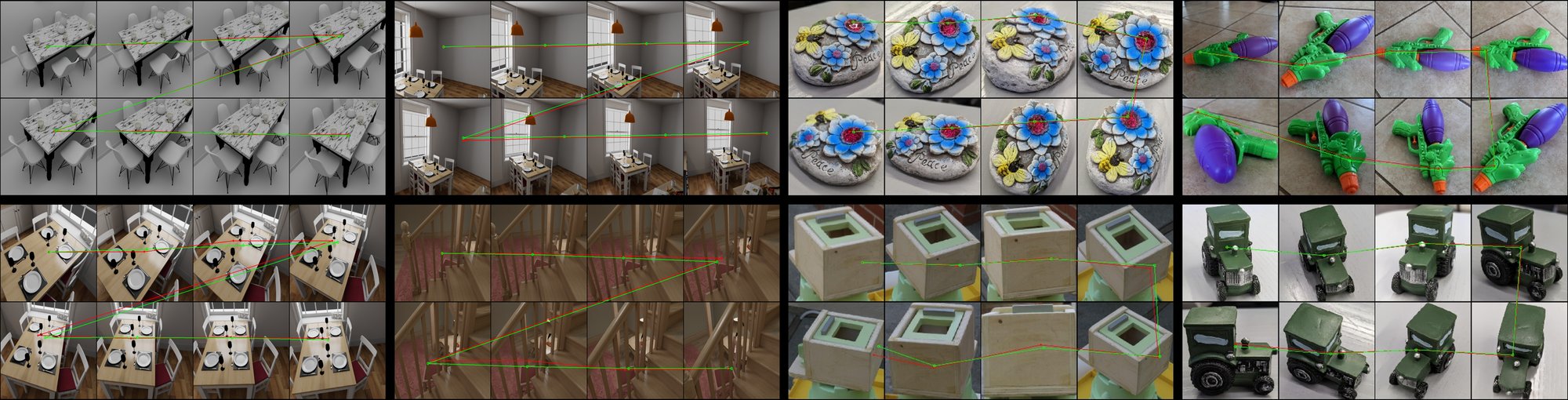}
    \end{minipage}
    \hfill
    \caption{
    \textbf{Qualitative comparison of correspondence from NAVI\cite{jampani2023navi} and NRGBD\cite{nrgbd_dataset_cvpr22}.} 
    }
    \label{fig:supply_tracks}
\end{figure*}

In the main paper, we evaluate the geometric consistency of our model on a zero-shot point tracking task. Here, we provide the formal definitions for the problem and evaluation metrics. We also present more qualitative examples of zero-shot correspondence in \cref{fig:supply_tracks}.
\paragraph{Formal Definitions.}
Given an image sequence comprising $V$ views, $\{I_v\}_{v=1}^V$. For each view $I_v$, we are provided with its corresponding intrinsic matrix ($K_v \in \mathbb{R}^{3 \times 3}$), depth map ($D_v$), camera Pose: $T_v \in SE(3)$, representing the camera-to-world transformation matrix.
The evaluation protocol is as follows:
\begin{enumerate}
    \item In the first view $I_1$, we sample $N$ starting points, $\{\mathbf{p}_{i,1}\}_{i=1}^N$, within valid depth regions. Here, $\mathbf{p}_{i,1} = (u_{i,1}, v_{i,1})$ denotes the pixel coordinates.
    \item The model's objective is to predict the corresponding locations of these points in all subsequent views $\{I_v\}_{v=2}^V$. For each starting point $\mathbf{p}_{i,1}$, the model generates a predicted trajectory $\mathcal{T}_i^{\text{pred}} = \{\mathbf{p}_{i,v}^{\text{pred}}\}_{v=1}^V$, where $\mathbf{p}_{i,v}^{\text{pred}} = (\hat{u}_{i,v}, \hat{v}_{i,v})$.
    \item We leverage the ground-truth depth and pose information to compute the true trajectory for each point, denoted as $\mathcal{T}_i^{\text{gt}} = \{(\mathbf{p}_{i,v}^{\text{gt}}, m_{i,v})\}_{v=1}^V$. Here, $\mathbf{p}_{i,v}^{\text{gt}}$ is the ground-truth 2D correspondence, and $m_{i,v} \in \{0, 1\}$ is a visibility mask, where $m_{i,v}=1$ if and only if point $i$ is visible in view $v$ (i.e., within the image bounds and not occluded).
\end{enumerate}

For a fair evaluation, all error metrics are computed exclusively over the set of {ground-truth visible points} (i.e., points where $m_{i,v}=1$).

\paragraph{Correspondence Extraction Methods.}
A key distinction in our evaluation lies in how correspondences are extracted by our method versus the baseline models. 
Baseline models like DINOv2, SAM, and MAE are fundamentally single-image encoders. To establish correspondences for these models, we first extract dense feature maps for each of the $V$ views independently. Subsequently, correspondences are found via a \textit{pairwise feature matching} process. Specifically, for each point in the first view, we find its nearest neighbor in the feature space of every other target view. This process is inherently pairwise; the match between view 1 and view $v$ does not leverage information from any other views in the sequence.
\themethod is designed with a native multi-view architecture. It processes all $V$ views simultaneously in a single forward pass. Correspondences are directly inferred from the model's internal \textit{cross-view attention maps}\cite{an2025zeroco}.

% --- 横跨双栏的三图并排 Figure ---
% 使用 figure* 环境使其横跨页面宽度
\begin{figure*}[t]
    \centering % 居中整个 figure
    \begin{subfigure}[b]{0.32\linewidth}
        \includegraphics[width=\linewidth]{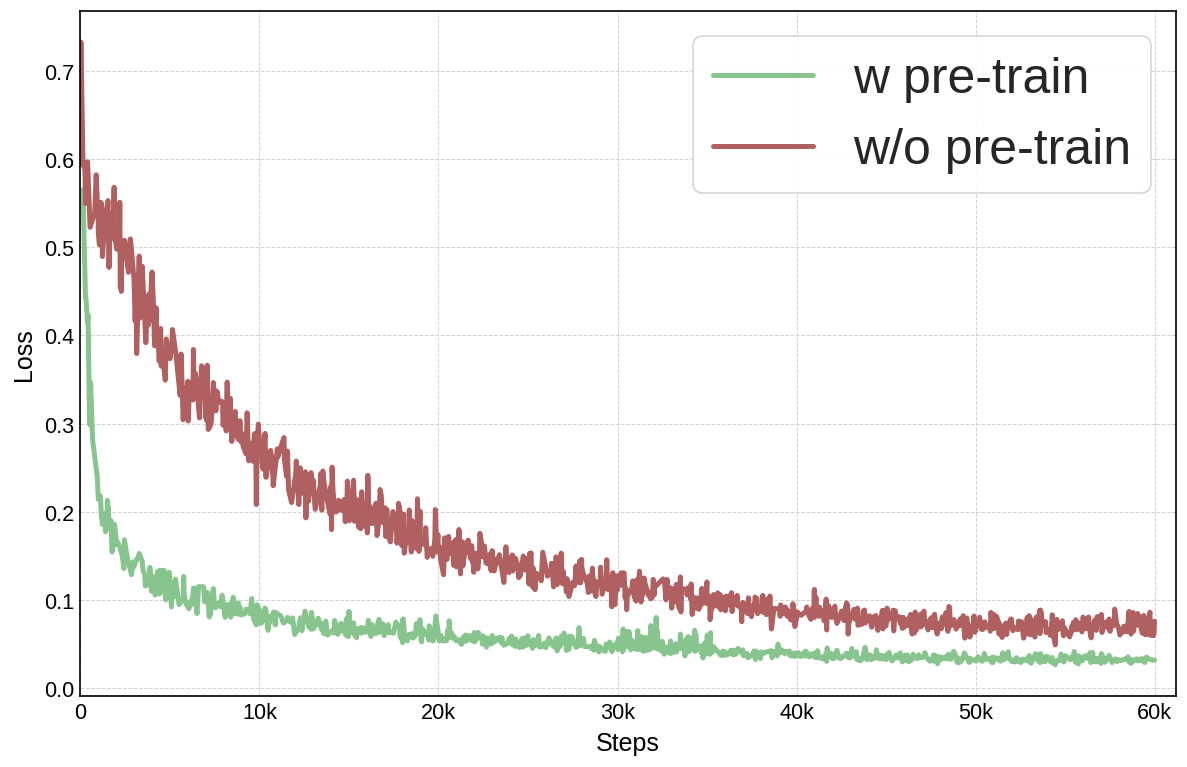}
        \subcaption{Training Loss}
        \label{fig:training_loss}
    \end{subfigure}
    \begin{subfigure}[b]{0.32\linewidth}
        \includegraphics[width=\linewidth]{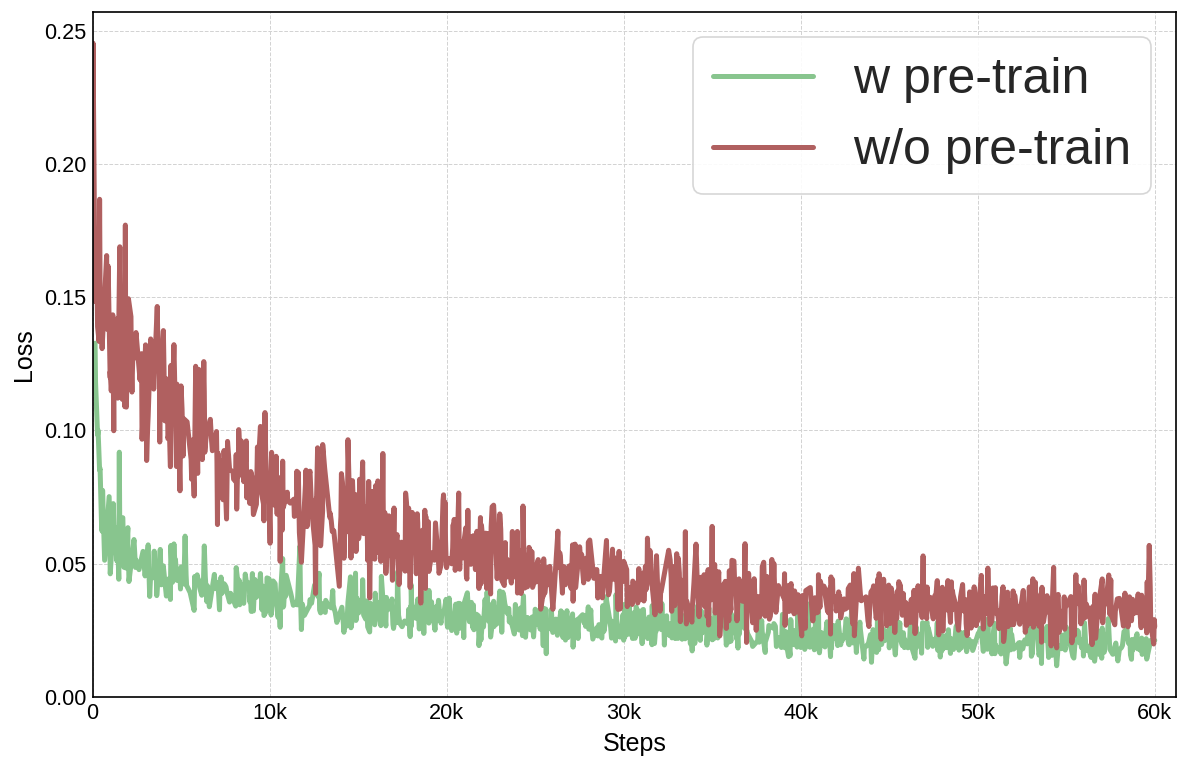}
        \subcaption{Pointmap Loss}
        \label{fig:pointmap_loss}
    \end{subfigure}
    \begin{subfigure}[b]{0.32\linewidth}
        \includegraphics[width=\linewidth]{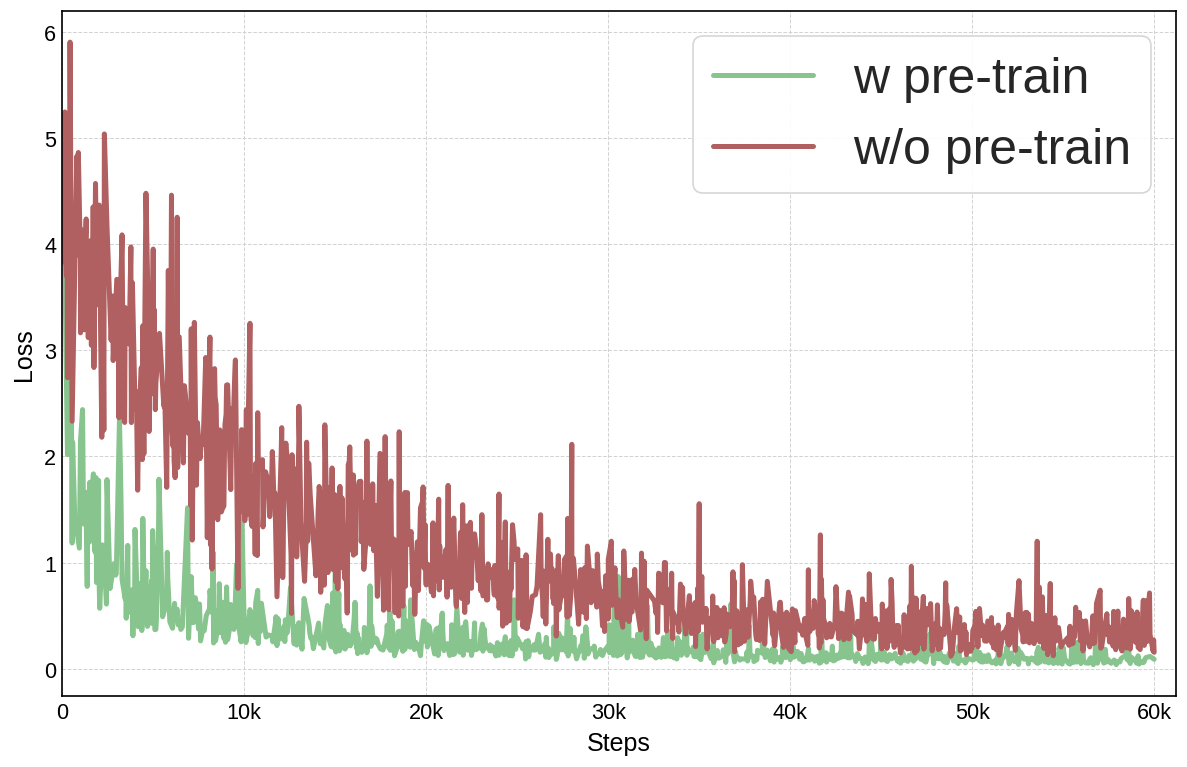}
        \subcaption{Camera Pose Loss}
        \label{fig:camera_loss}
    \end{subfigure}
    \caption{
        \textbf{Impact of Pre-training on Convergence Speed and Stability.} 
        This figure compares the loss curves when finetuned for 3D reconstruction with pre-trained weights (green) or with random initialization (red). The plots correspond to (a) overall reconstruction loss (b) loss for pointmap estimation and (c) loss for camera pose estimation.
        The pre-trained model demonstrates faster convergence and reaches a lower, more stable loss, validating the effectiveness of our pre-training strategy presented.
    }
    \label{fig:loss_ablation_pretrain}
\end{figure*}

\begin{table}[t!]
\centering
\small
\resizebox{\columnwidth}{!}{
\begin{tabular}{lccc}
\toprule
{Component} & {$\pi^3$}\cite{wang2025pi3} & {Baseline} & {Ours} \\
\midrule
Encoder & DINOv2-L\cite{oquab24dinov2:} & Frame-wise backbones & \themethod-L \\
\addlinespace % Adds a little extra vertical space
Decoder & 36 layers & 4 layers & 4 layers \\
\addlinespace
Specialized Heads & 5 layers & None & None \\
\bottomrule
\end{tabular}
}
\caption{
    \textbf{Comparison of Model Architectures.} 
    Our evaluation setup simplifies the original $\pi^3$ architecture. Our baseline experiments use various frame-wise ViT backbones (e.g., DINOv2, MAE, SAM). Crucially, all baselines and our method use an identical lightweight 4-layer decoder, which isolates the performance contribution of the encoder.
}
\label{tab:simplification_arch}
\end{table}

\begin{table}[t]
    \centering
    \footnotesize
    \resizebox{\columnwidth}{!}{
    \begin{tabular}{lcccccccc}
        \toprule
        \multirow{2}{*}{Method} & \multicolumn{8}{c}{Number of Input Frames ($N$)} \\
        \cmidrule(lr){2-9}
         & 1 & 4 & 8 & 10 & 20 & 50 & 100 & 200 \\
        \midrule
        DINOv3-B & {0.0106} & {0.0105} & {0.0108} & {0.0105} & {0.0129} & {0.0318} & {0.0594} & {0.1153} \\
        Muskie-B & 0.0113 & 0.0111 & 0.0114 & 0.0115 & 0.0157 & 0.0392 & 0.0987 & 0.2842 \\
        \bottomrule
    \end{tabular}
    }
    \caption{\textbf{Inference runtime comparison.} We measure the inference time (in seconds) across varying numbers of input frames. Both methods utilize PyTorch's optimized scaled dot product attention. Muskie demonstrates superior efficiency in standard low-frame settings ($N \le 10$) due to a streamlined architecture with lower constant overhead. At higher frame counts, the runtime increases moderately due to the global attention mechanism.}
    \label{tab:runtime_comparison}
\end{table}

\paragraph{2D Pixel Error.}
The 2D Pixel Error quantifies the discrepancy between the predicted point and the ground-truth correspondence directly in the image space.
For any point $i$ that is visible in view $v$ (i.e., $m_{i,v}=1$), its 2D pixel error, $e_{i,v}^{2D}$, is defined as the Euclidean distance between the predicted coordinates $\mathbf{p}_{i,v}^{\text{pred}}$ and the ground-truth coordinates $\mathbf{p}_{i,v}^{\text{gt}}$:
\begin{equation}
e_{i,v}^{2D} = \left\| \mathbf{p}_{i,v}^{\text{pred}} - \mathbf{p}_{i,v}^{\text{gt}} \right\|_2
\end{equation}

Based on this, we define two primary metrics:
\begin{itemize}
    \item {Average Trajectory Error (ATE in pixels)}: The mean 2D pixel error over all visible tracked points.
    \begin{equation}
    \text{ATE}^{2D} = \frac{\sum_{i=1}^N \sum_{v=1}^V m_{i,v} \cdot e_{i,v}^{2D}}{\sum_{i=1}^N \sum_{v=1}^V m_{i,v}}
    \end{equation}

    \item {Accuracy (Acc@k px)}: The percentage of visible points for which the 2D pixel error is less than a threshold of $k$ pixels.
    \begin{equation}
    \text{Acc@k px} = \frac{\sum_{i=1}^N \sum_{v=1}^V m_{i,v} \cdot \mathbb{I}(e_{i,v}^{2D} < k)}{\sum_{i=1}^N \sum_{v=1}^V m_{i,v}} \times 100\%
    \end{equation}
    where $\mathbb{I}(\cdot)$ is the indicator function.
\end{itemize}

\paragraph{3D Metric Error.}
The 3D Metric Error measures the positional deviation in 3D space that results from the 2D prediction error.
We first define an unprojection function $\pi^{-1}(\mathbf{p}, D, K)$, which lifts a 2D point $\mathbf{p}=(u,v)$ from a view into the camera's 3D coordinate system using the depth map $D$ and intrinsics $K$.
\begin{equation}
\mathbf{P} = \pi^{-1}(\mathbf{p}, D, K) = D(u,v) \cdot K^{-1} \cdot [u, v, 1]^T
\end{equation}
where $\mathbf{P}$ is the resulting 3D point coordinate.
For any point $i$ visible in view $v$ ($m_{i,v}=1$), we unproject both its ground-truth 2D coordinates and its predicted 2D coordinates into 3D space. We use the ground-truth depth map $D_v$ for both unprojections. This ensures that we are isolating the 3D error caused by the 2D tracking inaccuracy, rather than evaluating the model's own depth estimation capabilities.

\begin{itemize}
    \item {Ground-Truth 3D Point}: $\mathbf{P}_{i,v}^{\text{gt}} = \pi^{-1}(\mathbf{p}_{i,v}^{\text{gt}}, D_v, K_v)$
    \item {Predicted 3D Point}: $\mathbf{P}_{i,v}^{\text{pred}} = \pi^{-1}(\mathbf{p}_{i,v}^{\text{pred}}, D_v, K_v)$
\end{itemize}

The 3D metric error for this point, $e_{i,v}^{3D}$, is the Euclidean distance between these two 3D points:
\begin{equation}
e_{i,v}^{3D} = \left\| \mathbf{P}_{i,v}^{\text{pred}} - \mathbf{P}_{i,v}^{\text{gt}} \right\|_2
\end{equation}

This leads to the corresponding 3D metrics:
\begin{itemize}
    \item {Average Trajectory Error (ATE in cm)}: The mean 3D metric error over all visible tracked points, with units converted from meters to centimeters.
    \begin{equation}
    \text{ATE}^{3D} = \left( \frac{\sum_{i=1}^N \sum_{v=1}^V m_{i,v} \cdot e_{i,v}^{3D}}{\sum_{i=1}^N \sum_{v=1}^V m_{i,v}} \right) \times 100
    \end{equation}

    \item {Accuracy (Acc@k cm)}: The percentage of visible points for which the 3D metric error is less than a threshold of $k$ centimeters.
    \begin{equation}
    \text{Acc@k cm} = \frac{\sum_{i=1}^N \sum_{v=1}^V m_{i,v} \cdot \mathbb{I}(e_{i,v}^{3D} < k/100)}{\sum_{i=1}^N \sum_{v=1}^V m_{i,v}} \times 100\%
    \end{equation}
\end{itemize}

\begin{table*}[htbp]
\centering
\small
\setlength{\tabcolsep}{2pt} 
\resizebox{\textwidth}{!}{%
\begin{tabular}{@{}l | ccccccc | ccccc@{}}
\toprule
\multirow{2}{*}{Method} & \multicolumn{7}{c|}{Pixel Space (px)} & \multicolumn{5}{c}{3D Space (cm)} \\
\cmidrule(lr){2-8} \cmidrule(lr){9-13}
 & ATE$_\text{2D}$ $\downarrow$ & Acc@1px$\uparrow$ & Acc@2px$\uparrow$ & Acc@5px$\uparrow$ & Acc@10px$\uparrow$ & Acc@25px$\uparrow$ & Acc@50px$\uparrow$ & ATE$_\text{3D}$ $\downarrow$ & Acc@1cm$\uparrow$ & Acc@2cm$\uparrow$ & Acc@5cm$\uparrow$ & Acc@10cm$\uparrow$ \\
\midrule
DINOv2-B\cite{oquab24dinov2:}         & 56.46 & 13.01 & 13.53 & 16.66 & 25.55 & 50.15 & 68.68 & 4.43 & 53.72 & 68.98 & 83.33 & 88.34 \\
DINOv2-L\cite{oquab24dinov2:}         & 53.31 & 13.02 & 13.56 & 16.94 & 26.29 & 51.37 & 69.87 & 4.24 & 54.72 & 69.97 & 84.23 & 88.86 \\
DINOv3-B\cite{simeoni2025dinov3}      & 42.95 & 13.07 & 13.59 & 17.12 & 27.25 & 54.65 & 74.12 & 3.92 & 57.74 & 73.55 & 86.44 & 89.53 \\
DINOv3-L\cite{simeoni2025dinov3}      & \underline{41.15} & 13.02 & 13.53 & 17.23 & 27.95 & \underline{57.11} & \underline{76.14} & 3.76 & 60.15 & 74.89 & 87.11 & 90.00 \\
% \midrule
SAM\cite{sam}                         & 69.54 & 12.98 & 13.32 & 15.68 & 22.28 & 41.36 & 58.16 & 4.56 & 46.65 & 62.58 & 81.66 & 88.73 \\
MAE\cite{he21masked}                  & 81.24 & 12.91 & 13.10 & 14.63 & 19.11 & 33.01 & 48.51 & 5.25 & 37.75 & 54.73 & 78.19 & 86.24 \\
DIFT\cite{dift}                       & 60.44 & 13.01 & 13.56 & 16.97 & 26.46 & 49.55 & 63.75 & 4.93 & 51.00 & 64.36 & 81.21 & 86.83 \\
SigLIP\cite{siglip}                   & 111.66 & 12.86 & 12.94 & 13.66 & 15.71 & 24.21 & 37.64 & 7.79 & 28.37 & 43.46 & 68.41 & 79.35 \\
CLIP\cite{radford21learning}          & 189.22 & 12.85 & 12.87 & 12.97 & 13.35 & 15.30 & 19.65 & 13.86 & 17.22 & 23.84 & 41.10 & 55.03 \\
CroCo\cite{weinzaepfel2023crocov2}    & 56.79 & 13.15 & \underline{14.07} & \underline{19.77} & \underline{32.58} & 54.77 & 69.74 & 4.42 & 58.13 & 72.15 & 84.66 & 87.39 \\
% \midrule
\themethod-B                          & 42.54 & \underline{13.15} & 14.03 & 19.48 & 31.95 & 56.17 & 73.86 & \underline{2.93} & \underline{60.92} & \underline{75.90} & \underline{89.17} & \underline{93.30} \\
\themethod-L                          & \textbf{29.42} & \textbf{13.28} & \textbf{14.53} & \textbf{22.29} & \textbf{39.21} & \textbf{68.69} & \textbf{83.98} & \textbf{2.38} & \textbf{71.03} & \textbf{82.74} & \textbf{91.87} & \textbf{94.32} \\
\bottomrule
\end{tabular}%
}
\caption{
    \textbf{Quantitative results for zero-shot point tracking on the NAVI \cite{jampani2023navi} dataset across 8 views.} 
    We compare our \themethod models against a comprehensive set of foundational models. Best results are in \textbf{bold}, and \underline{underlined} results indicate the second best.
}
\label{tab:navi_8_view_tracking}
\end{table*}
\begin{table*}[htbp]
\centering
\small
\setlength{\tabcolsep}{2pt} 
\resizebox{\textwidth}{!}{%
\begin{tabular}{@{}l | ccccccc | ccccc@{}}
\toprule
\multirow{2}{*}{Method} & \multicolumn{7}{c|}{Pixel Space (px)} & \multicolumn{5}{c}{3D Space (cm)} \\
\cmidrule(lr){2-8} \cmidrule(lr){9-13}
 & ATE$_\text{2D}$ $\downarrow$ & Acc@1px$\uparrow$ & Acc@2px$\uparrow$ & Acc@5px$\uparrow$ & Acc@10px$\uparrow$ & Acc@25px$\uparrow$ & Acc@50px$\uparrow$ & ATE$_\text{3D}$ $\downarrow$ & Acc@1cm$\uparrow$ & Acc@2cm$\uparrow$ & Acc@5cm$\uparrow$ & Acc@10cm$\uparrow$ \\
\midrule
DINOv2-B\cite{oquab24dinov2:}         & 72.41 & 20.76 & 21.19 & 23.17 & 29.32 & 45.87 & 60.31 & 49.88 & 24.57 & 27.31 & 35.00 & 45.71 \\
DINOv2-L\cite{oquab24dinov2:}         & 72.11 & 20.85 & 21.26 & 23.42 & 29.22 & 44.16 & 59.17 & 49.09 & 24.55 & 27.13 & 34.29 & 44.64 \\
DINOv3-B\cite{simeoni2025dinov3}      & 64.07 & 20.83 & 21.17 & 24.41 & 32.08 & 53.74 & 66.77 & 46.15 & 25.08 & 28.98 & 40.22 & 51.09 \\
DINOv3-L\cite{simeoni2025dinov3}      & 55.78 & 20.85 & 21.26 & 24.66 & 33.54 & 56.06 & 70.32 & 42.58 & 25.27 & 29.49 & 41.69 & 52.87 \\
% \midrule
SAM\cite{sam}                         & 76.12 & 20.77 & 21.01 & 23.80 & 31.16 & 47.85 & 60.31 & 49.85 & 24.01 & 27.34 & 36.56 & 46.05 \\
MAE\cite{he21masked}                  & 104.94 & 20.70 & 20.85 & 22.07 & 25.18 & 33.48 & 44.10 & 62.70 & 22.81 & 24.01 & 28.51 & 34.89 \\
DIFT\cite{dift}                       & 69.82 & 20.88 & 21.55 & 26.44 & 39.21 & 59.86 & 68.15 & 50.56 & 25.76 & 31.65 & 44.28 & 54.62 \\
SigLIP\cite{siglip}                   & 142.96 & 20.68 & 20.74 & 21.04 & 21.78 & 26.16 & 34.02 & 81.44 & 22.72 & 23.04 & 24.88 & 28.75 \\
CLIP\cite{radford21learning}          & 203.18 & 20.67 & 20.67 & 20.83 & 21.04 & 22.21 & 25.09 & 102.45 & 22.30 & 22.47 & 22.95 & 24.34 \\
CroCo\cite{weinzaepfel2023crocov2}    & \underline{25.92} & 21.19 & \underline{23.08} & \textbf{34.20} & \textbf{55.93} & \textbf{80.09} & \textbf{89.27} & \underline{29.84} & \underline{28.19} & \textbf{38.16} & \textbf{58.02} & \textbf{72.08} \\
% \midrule
\themethod-B                          & 33.77 & \underline{21.24} & 22.77 & 31.41 & 47.67 & 68.37 & 81.60 & 31.67 & 27.18 & 35.03 & 51.14 & 63.92 \\
\themethod-L                          & \textbf{24.68} & \textbf{21.33} & \textbf{23.15} & \underline{33.83} & \underline{52.96} & \underline{76.49} & \underline{87.90} & \textbf{27.82} & \textbf{28.44} & \underline{37.52} & \underline{56.60} & \underline{70.84} \\
\bottomrule
\end{tabular}%
}
\caption{
    \textbf{Quantitative results for zero-shot point tracking on the ScanNet\cite{dai2017scannet} dataset across 8 views.} 
    We compare our \themethod models against a comprehensive set of foundational models. Best results are in \textbf{bold}, and \underline{underlined} results indicate the second best.
}
\label{tab:scannet_8_view_tracking}
\end{table*}
\subsection{3D Reconstruction}
\begin{figure*}[t]
    \centering % 居中整个 figure
    \begin{subfigure}[b]{0.18\linewidth}
        \includegraphics[width=\linewidth]{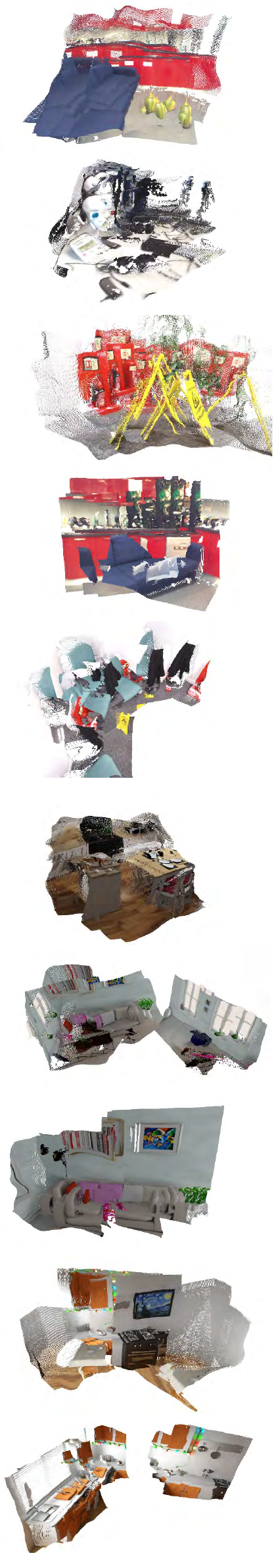}
        \subcaption{MAE}
    \end{subfigure}
    \begin{subfigure}[b]{0.1612\linewidth}
        \includegraphics[width=\linewidth]{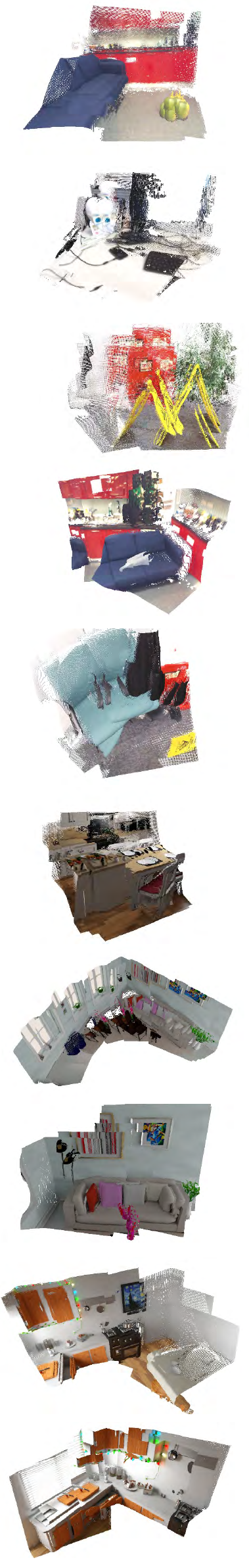}
        \subcaption{DINOv2}
    \end{subfigure}
    \begin{subfigure}[b]{0.1715\linewidth}
        \includegraphics[width=\linewidth]{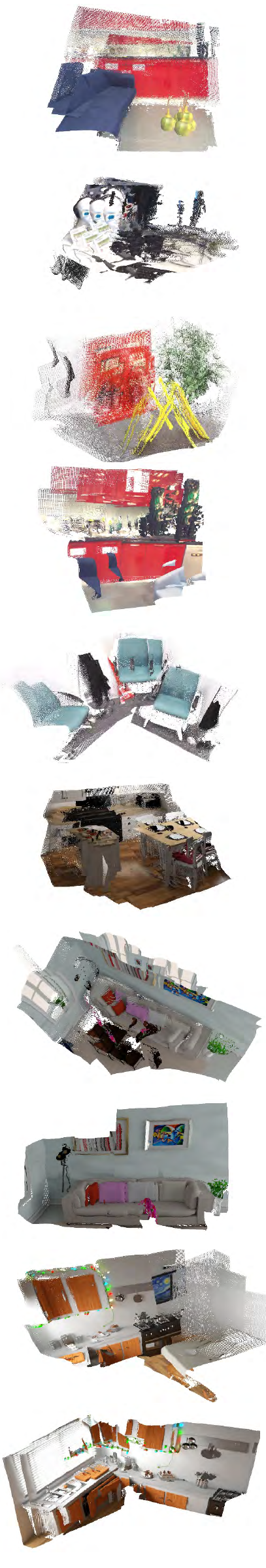}
        \subcaption{DINOv3}
    \end{subfigure}
    \begin{subfigure}[b]{0.18\linewidth}
        \includegraphics[width=\linewidth]{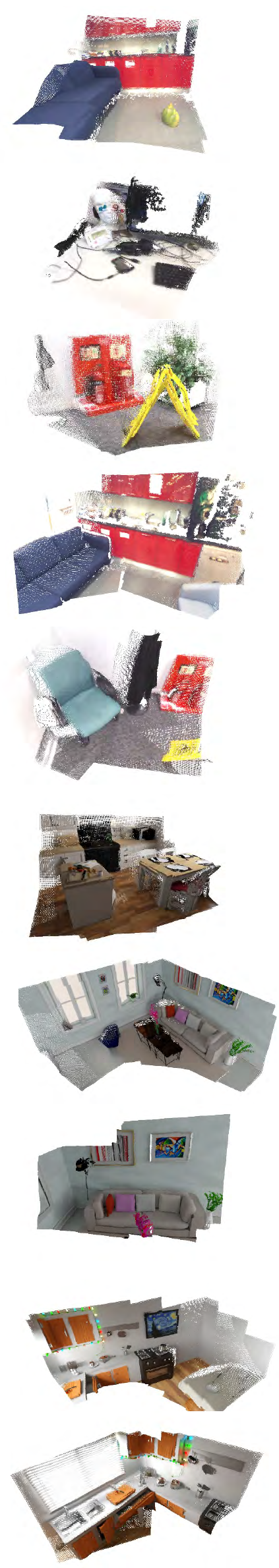}
        \subcaption{Ours}
    \end{subfigure}
    \begin{subfigure}[b]{0.1735\linewidth}
        \includegraphics[width=\linewidth]{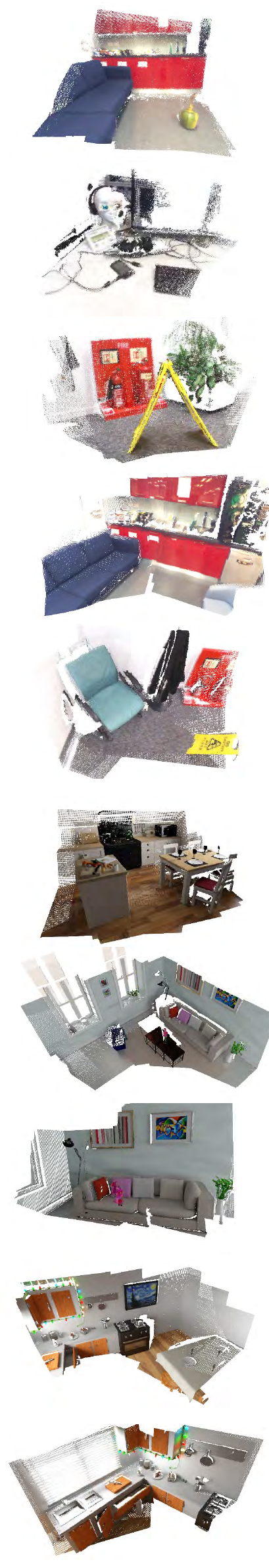}
        \subcaption{Ground Truth}
    \end{subfigure}
    \caption{
        \textbf{Qualitative comparison of predicted pointmaps for NRGBD\cite{nrgbd_dataset_cvpr22} and 7Scenes\cite{7scenes}.} 
    }
    \label{fig:supply_pointmap}
\end{figure*}

\paragraph{Finetune Setting.}
To ensure a fair and controlled comparison, we designed our experimental setup to specifically isolate the contribution of the feature encoder, as detailed in \cref{tab:simplification_arch}.
% Note that the original $\pi^3$ architecture employs a powerful and computationally intensive decoder (36 layers plus specialized heads), which was prohibitive for our research setting.
We present more qualitative examples of reconstructed pointmaps in \cref{fig:supply_pointmap}. 

\paragraph{CroCo as backbone.}
Due to space constraints in the main manuscript, we focused primarily on comparisons with standard frame-wise backbones.
To provide a deeper analysis of multi-view capabilities, we present additional comparisons with CroCo\cite{weinzaepfel2023crocov2} in this section. CroCo is a representative method that learns cross-view consistency through pairwise masked image modeling.
When utilizing CroCo\cite{weinzaepfel2023crocov2} as a feature backbone, we explore two distinct strategies to extract latent representations from multi-view inputs.
\textit{The first setting is to treat CroCo solely as a feature extractor by utilizing only its encoder.}
Given a sequence of multi-view images, we reshape the batch to process each frame independently.
Each image is passed through the CroCo encoder, and the output from the final encoder block is extracted as the latent representation. This approach aligns with standard frame-wise backbones like DINO\cite{caron2021dino}, offering efficiency but lacking explicit cross-view interaction during feature extraction.
\textit{The other strategy is to leverage its cross-view completion capability, where we employ a reference-based pairwise strategy.}
We select the first view in the sequence as the reference source. For every target view in the batch (including the first view), we construct a source-target pair and pass them through the full encoder-decoder pipeline. Specifically, both the reference and target images are encoded, and their features are then fed into the decoder, where cross-attention mechanisms allow the reference view to refine the target view's representation. We extract the output from the decoder blocks as the final, geometrically enriched features for the target view. While this method introduces cross-view context, it incurs a higher computational cost due to the pairwise processing of 
$N$ views against a fixed reference.
It is worth noting that the CroCo architecture consists of a ViT-Large encoder coupled with a ViT-Base decoder, resulting in a higher total parameter count compared to other baselines.
This difference introduces a disparity in model capacity, theoretically biasing the comparison in favor of CroCo.
Despite this unfair advantage in model capacity favoring CroCo, our method still achieves superior performance, highlighting the efficiency of our proposed approach, as shown in \cref{tab:croco_pose_comparison,tab:croco_pointmap_comparison}.

\begin{table}[t!]
  \centering
  \resizebox{\columnwidth}{!}{
  \setlength{\tabcolsep}{4pt} % 可选：如果表格太宽，稍微减小列间距
  \begin{tabular}{lcccccc}
    \toprule
    \multirow{2}{*}{{Method}} & \multicolumn{3}{c}{{NRGBD\cite{nrgbd_dataset_cvpr22}}} & \multicolumn{3}{c}{{7Scenes\cite{7scenes}}} \\
    \cmidrule(lr){2-4} \cmidrule(lr){5-7}
    & {Acc} & {Comp} & {$||\mathcal{L}_1||$} & {Acc} & {Comp} & {$||\mathcal{L}_1||$} \\
    \midrule
    CroCo (encoder) & 0.0647 & 0.1127 & 0.1179 & 0.0429 & 0.0526 & 0.0864 \\
    CroCo (decoder) & 0.0541 & 0.0889 & 0.0823 & 0.0311 & 0.0321 & 0.0491 \\
    \themethod-B    & 0.0585 & 0.0896 & 0.0897 & 0.0391 & 0.0405 & 0.0508 \\
    \themethod-L    & \textbf{0.0460} & \textbf{0.0765} & \textbf{0.0667} & \textbf{0.0249} & \textbf{0.0283} & \textbf{0.0346} \\
    \bottomrule
  \end{tabular}
  }
  \caption{Performance comparison on pointmap across NRGBD \cite{nrgbd_dataset_cvpr22} and 7Scenes\cite{7scenes} datasets.}
  \label{tab:croco_pointmap_comparison}
\end{table}

\begin{table}[t!]
  \centering
  \resizebox{\columnwidth}{!}{
  \setlength{\tabcolsep}{3pt} % 稍微减小列间距以适应页面
  \begin{tabular}{lcccccc}
    \toprule
    \multirow{2}{*}{{Method}} & \multicolumn{3}{c}{{NRGBD\cite{nrgbd_dataset_cvpr22}}} & \multicolumn{3}{c}{{7Scenes\cite{7scenes}}} \\
    \cmidrule(lr){2-4} \cmidrule(lr){5-7}
    & {AUC$_5$} & {AUC$_{15}$} & {AUC$_{30}$} & {AUC$_5$} & {AUC$_{15}$} & {AUC$_{30}$} \\
    \midrule
    CroCo (encoder) & 0.63 & 7.12 & 16.92 & 0.23 & 2.86 & 9.98 \\
    CroCo (decoder) & 7.03 & 28.56 & 45.45 & 2.96 & 16.34 & 32.60 \\
    \themethod-B    & 3.69 & 25.04 & 45.43 & 1.41 & 13.81 & 31.23 \\
    \themethod-L    & \textbf{19.46} & \textbf{50.96} & \textbf{67.45} & \textbf{6.10} & \textbf{26.93} & \textbf{47.35} \\
    \bottomrule
  \end{tabular}
  }
  \caption{Camera pose estimation performance (AUC) comparison on NRGBD \cite{nrgbd_dataset_cvpr22} and 7Scenes\cite{7scenes} datasets. }
  \label{tab:croco_pose_comparison}
\end{table}

\subsection{Ablation Studies}

\paragraph{Pretraining or Architecture.}
As shown in Fig.~\ref{fig:loss_ablation_pretrain}, using \themethod pretrained weights as a starting point leads to accelerated convergence speeds and superior training stability compared to random initialization, characterized by reduced loss variance and smoother optimization trajectories.

% \paragraph{Confidence Loss} \TODO{drop it.}

\begin{figure}[t!]
    \centering % 居中整个 figure
    \includegraphics[width=.9\linewidth]{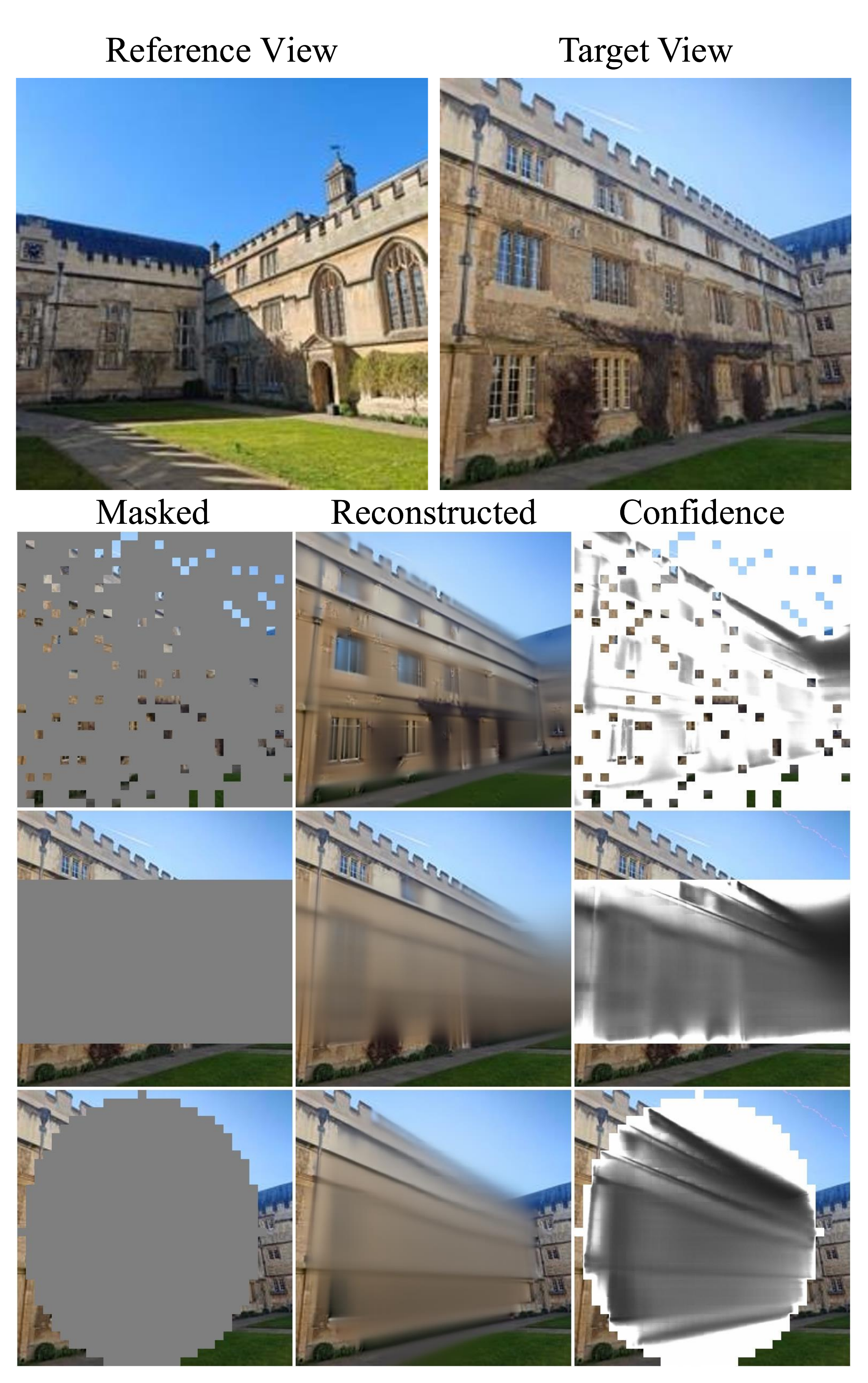}
    \caption{
        Reconstruction fails when there is no overlap between the reference view and the view to be reconstructed. Source images are from \cite{wang2025vggt}.
    }
    \label{fig:failure_case}
\end{figure}
\subsection{Pre-training}

\begin{figure*}[t]
    \centering % 居中整个 figure
    \begin{subfigure}[b]{0.3\linewidth}
        \includegraphics[width=\linewidth]{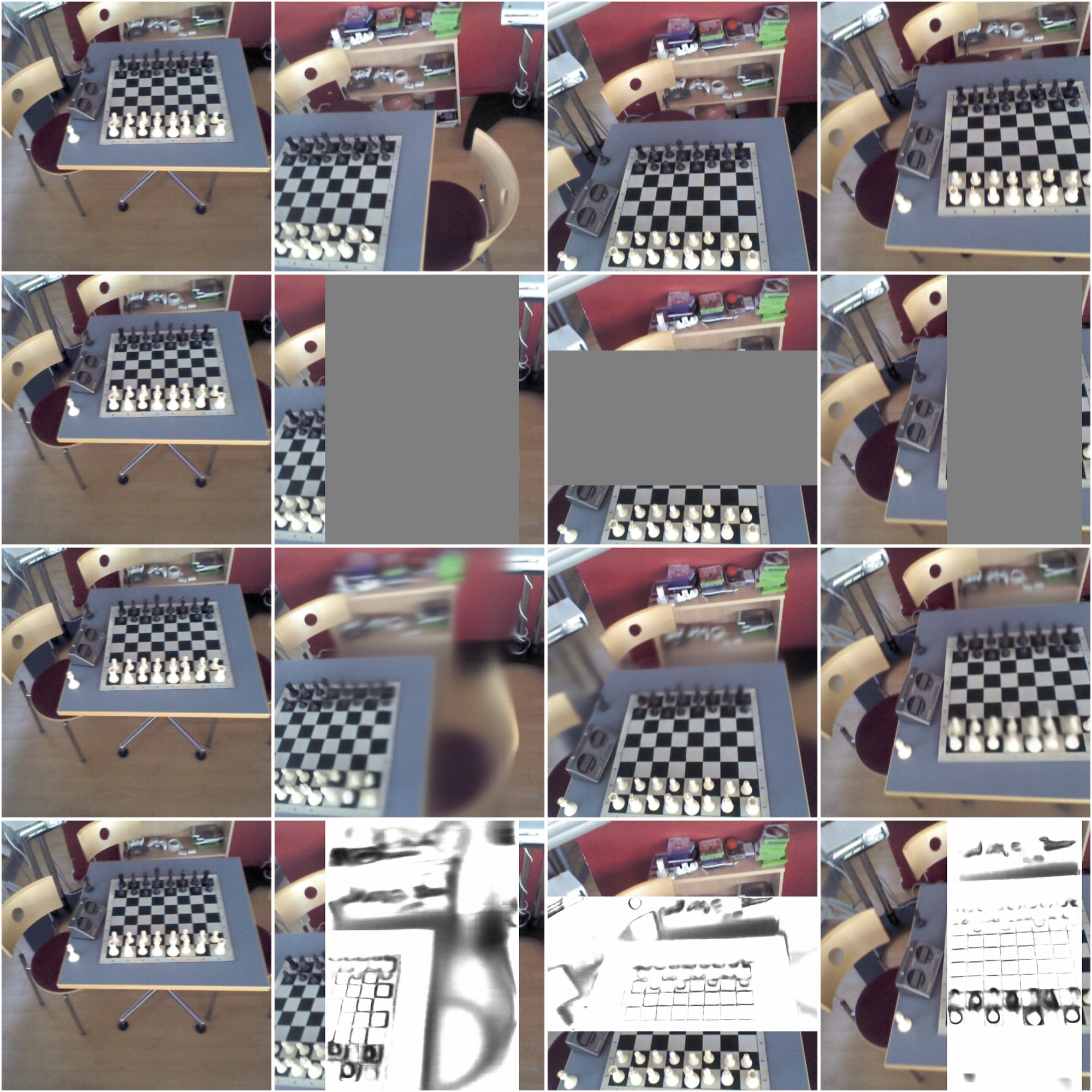}
    \end{subfigure}
    \hspace{0.01\linewidth}
    \begin{subfigure}[b]{0.6\linewidth}
        \includegraphics[width=\linewidth]{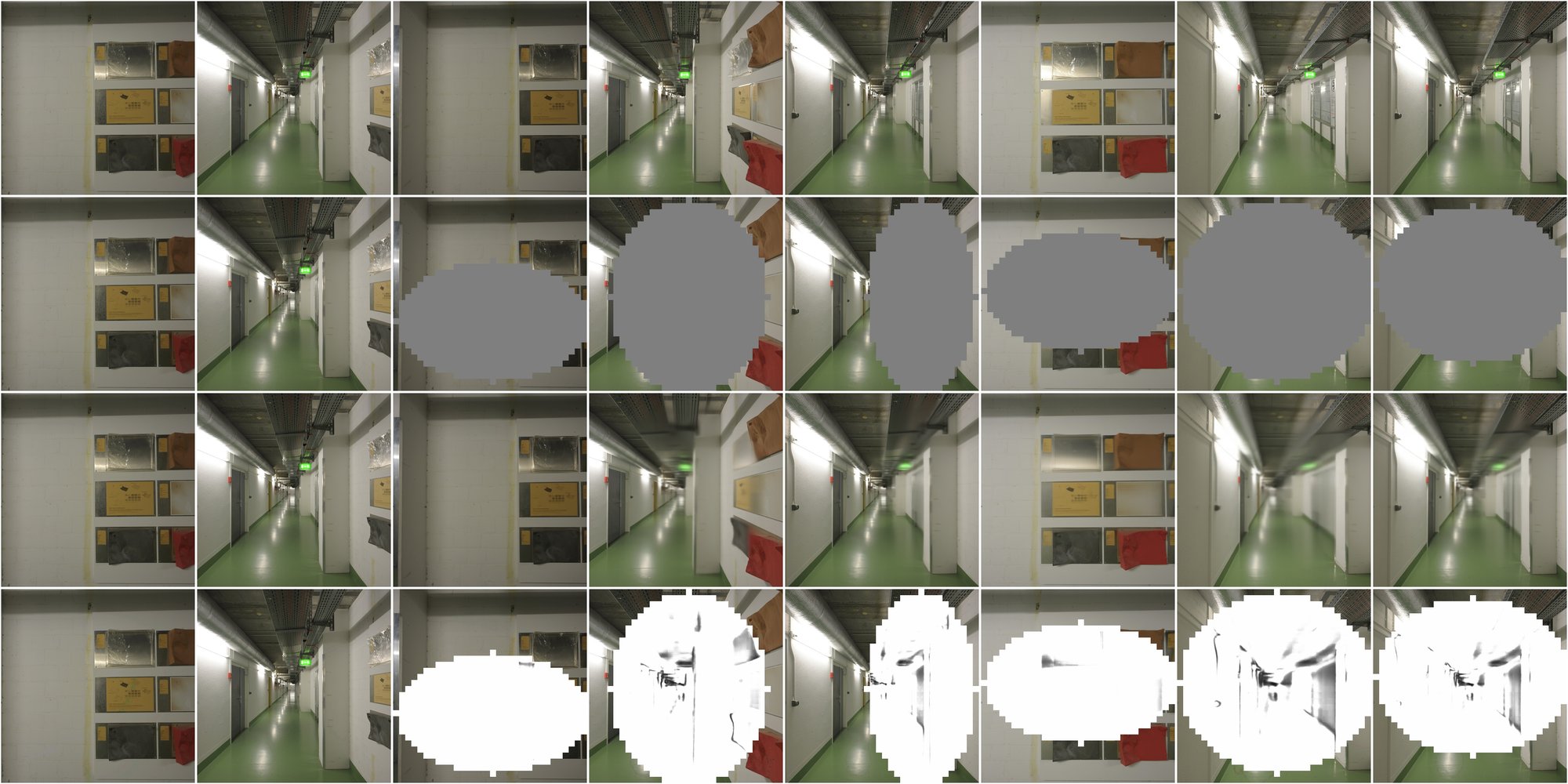}
    \end{subfigure}
    
    \vspace{0.01\linewidth}
    
    \begin{subfigure}[b]{0.3\linewidth}
        \includegraphics[width=\linewidth]{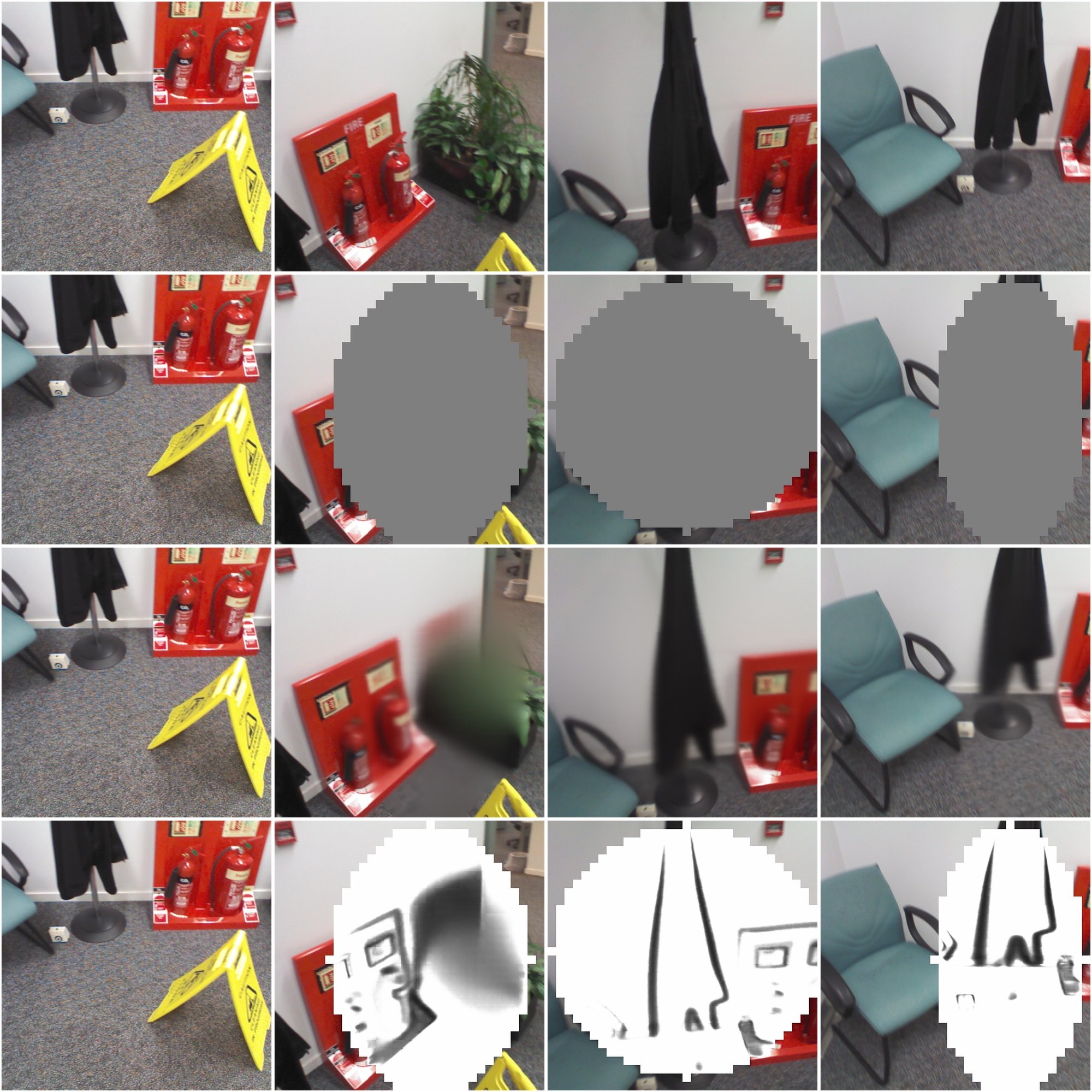}
    \end{subfigure}
    \hspace{0.01\linewidth}
    \begin{subfigure}[b]{0.6\linewidth}
        \includegraphics[width=\linewidth]{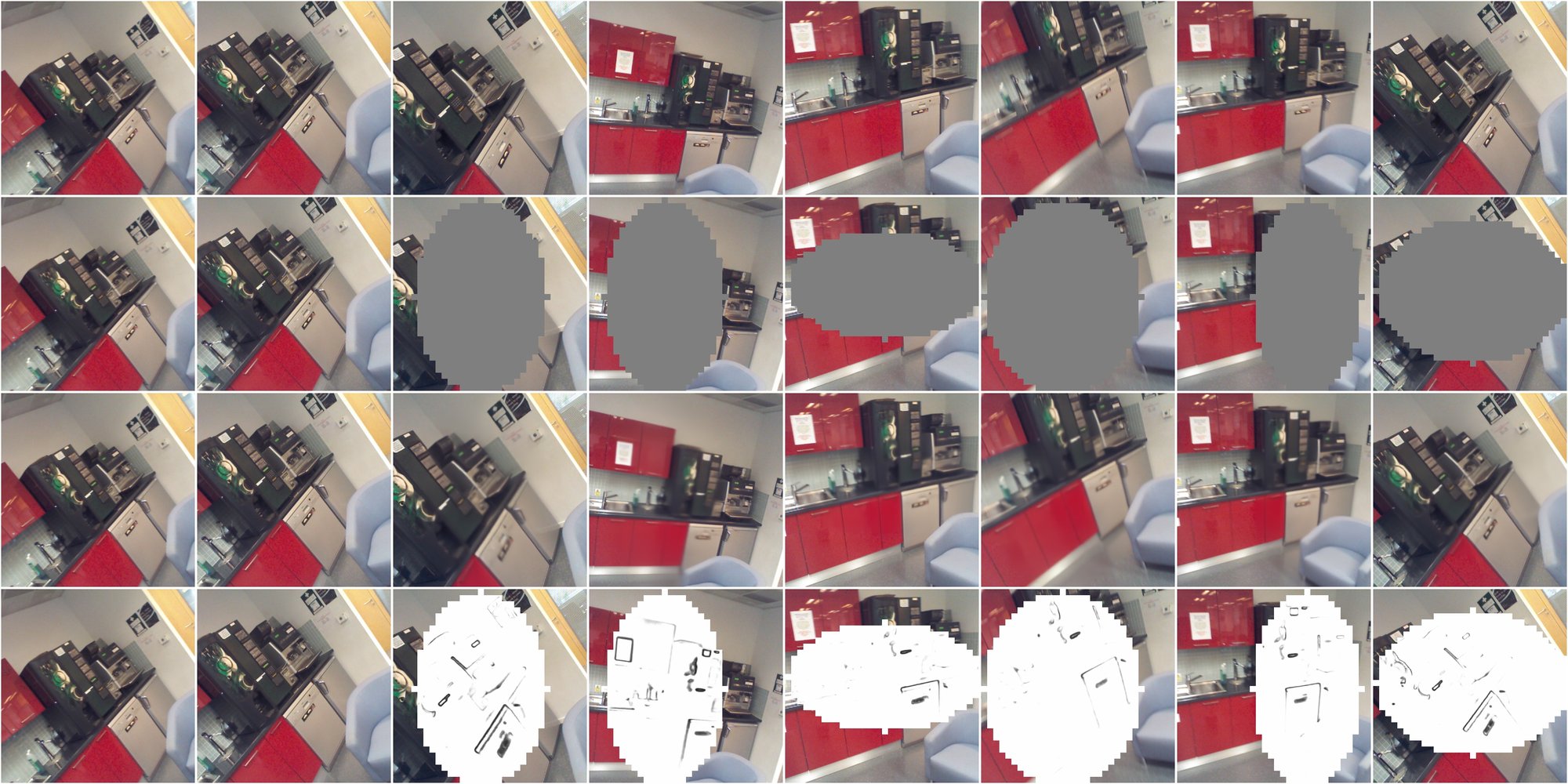}
    \end{subfigure}
    
    \vspace{0.01\linewidth}
    
    \begin{subfigure}[b]{0.3\linewidth}
        \includegraphics[width=\linewidth]{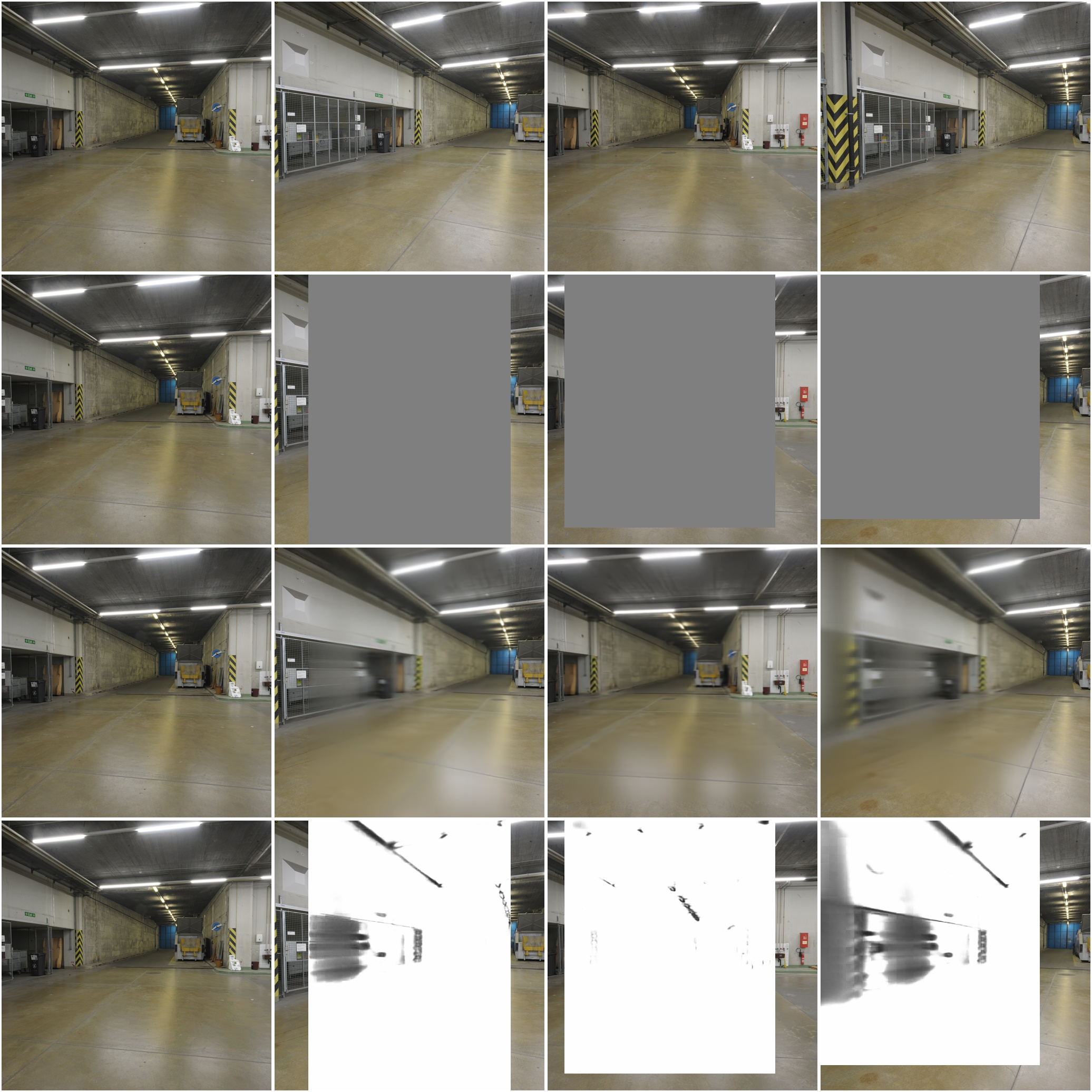}
    \end{subfigure}
    \hspace{0.01\linewidth}
    \begin{subfigure}[b]{0.6\linewidth}
        \includegraphics[width=\linewidth]{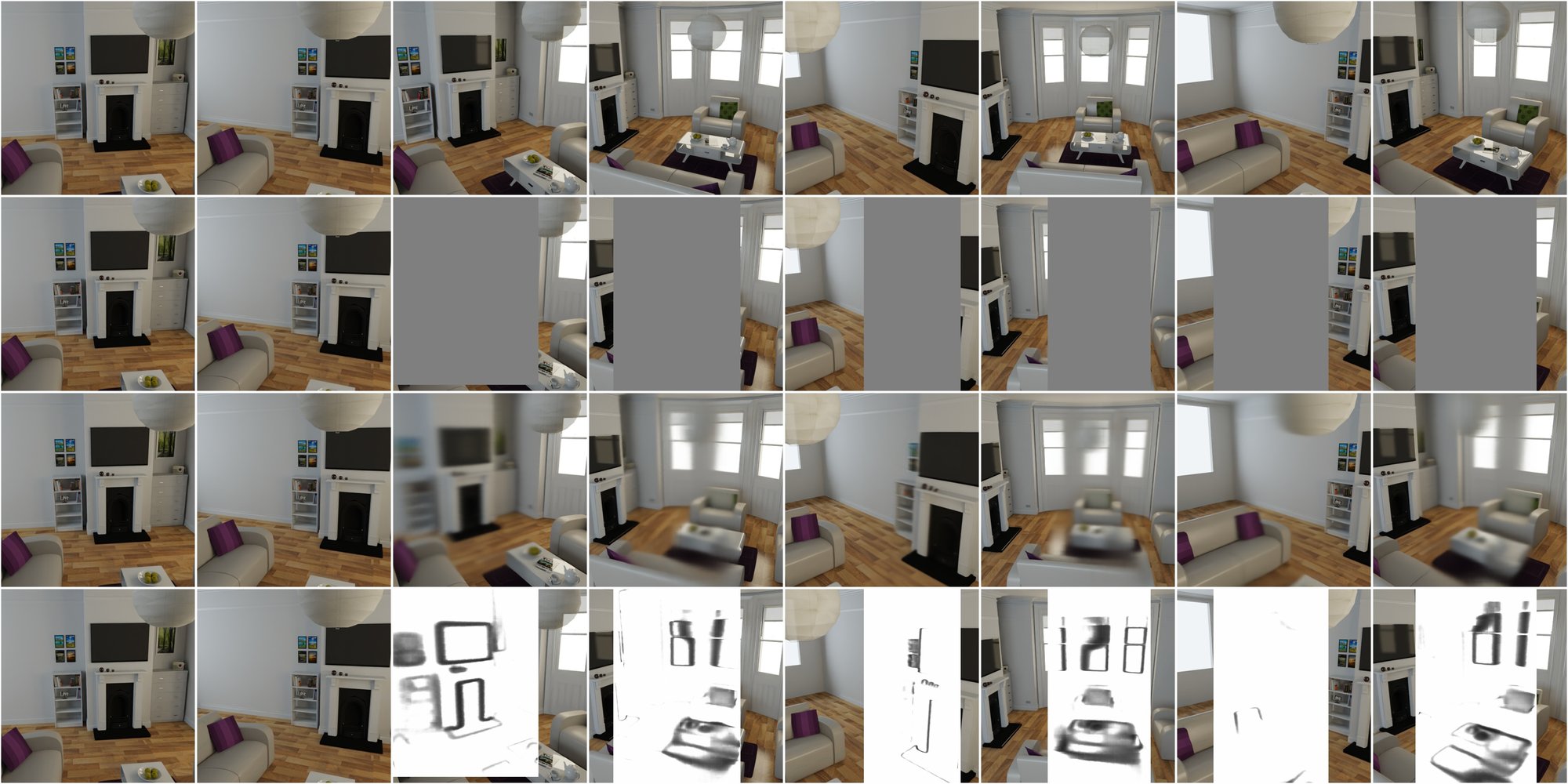}
    \end{subfigure}

    \vspace{0.01\linewidth}
    
    \begin{subfigure}[b]{0.3\linewidth}
        \includegraphics[width=\linewidth]{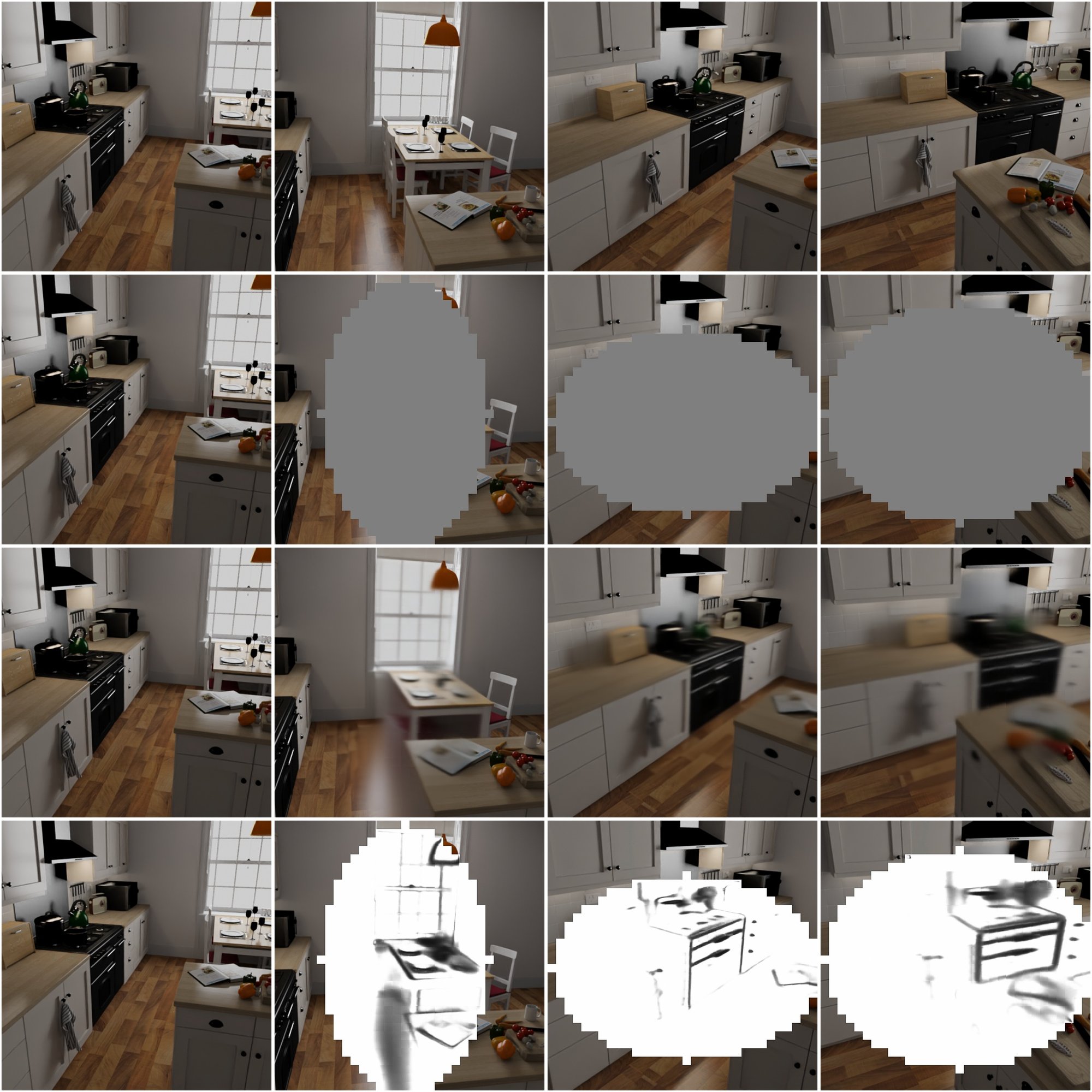}
    \end{subfigure}
    \hspace{0.01\linewidth}
    \begin{subfigure}[b]{0.6\linewidth}
        \includegraphics[width=\linewidth]{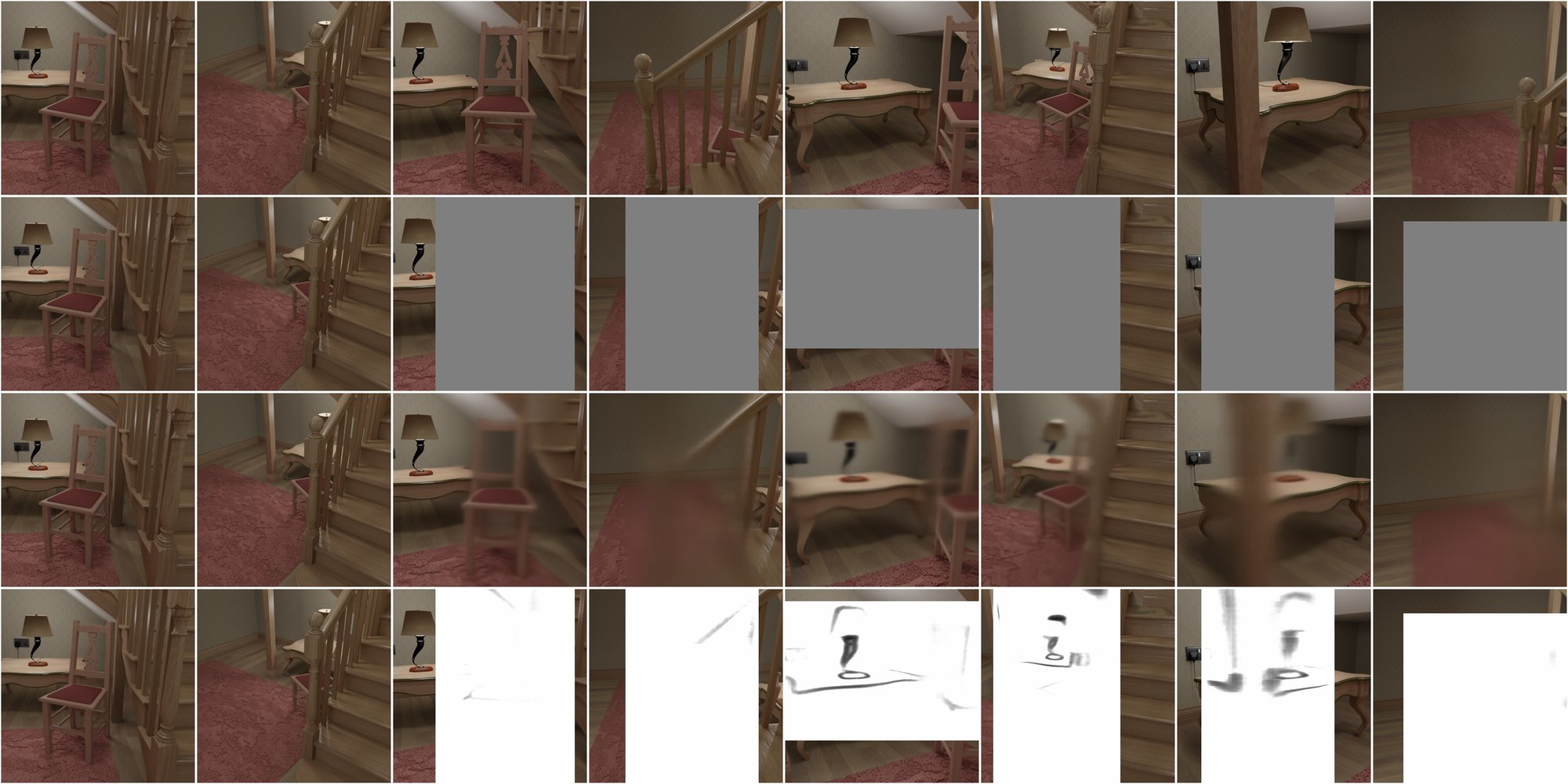}
    \end{subfigure}
    \caption{
        \textbf{Additional examples of reconstruction from masked multi-view images\cite{eth3d,nrgbd_dataset_cvpr22,7scenes}. } 
    }
    \label{fig:more_masking_examples}
\end{figure*}

\paragraph{More results}
In this section, we further present additional cases where \themethod reconstructs complete images from masked multi-view inputs, as shown in \cref{fig:more_masking_examples}. These examples were not used for training.

\paragraph{Non-overlapping Case}
When the input view pair contains no overlapping regions, \themethod's multi-view capability effectively degenerates into single-view masked image modeling. As shown in \cref{fig:failure_case}, with randomly scattered masks, the network can still exploit the remaining sparse visible patches to reconstruct the target, exhibiting behavior similar to MAE\cite{he21masked}. In contrast, when large aggregated masks are applied, most intra-view structural cues are removed. Without cross-view information to compensate for the missing content, the model produces blurred reconstructions, and the low confidence indicates the absence of usable multi-view cues.

\begin{figure}[t!]
\centering
\begin{tabular}{ >{\centering\arraybackslash}m{0mm}  m{0.85\linewidth} }

% ===== Inputs =====
\rotatebox{90}{\small Inputs} &
\includegraphics[width=\linewidth]{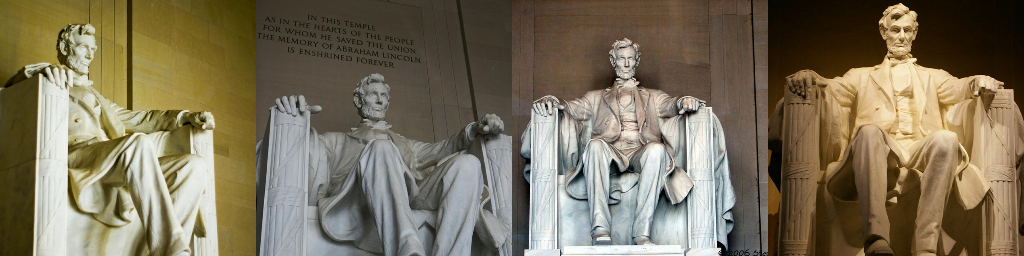} \\
[3mm]

% ===== Layer 2 =====
\rotatebox{90}{\small Layer 4} &
\includegraphics[width=\linewidth]{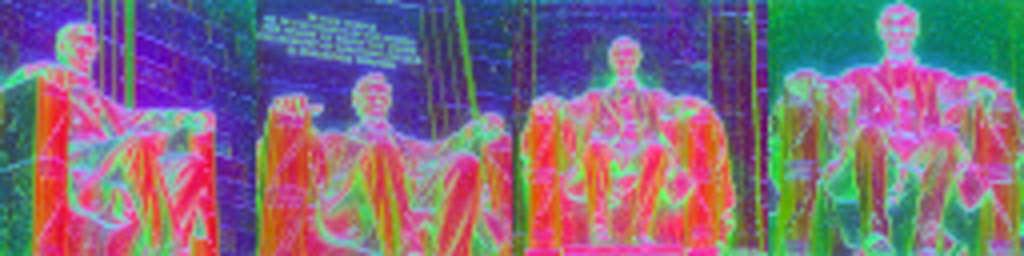} \\
[0mm]

% ===== Layer 4 =====
\rotatebox{90}{\small Layer 8} &
\includegraphics[width=\linewidth]{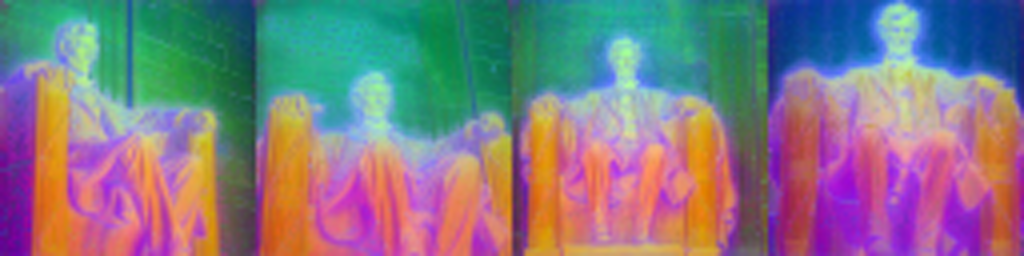} \\
[0mm]

% ===== Layer 6 =====
\rotatebox{90}{\small Layer 12} &
\includegraphics[width=\linewidth]{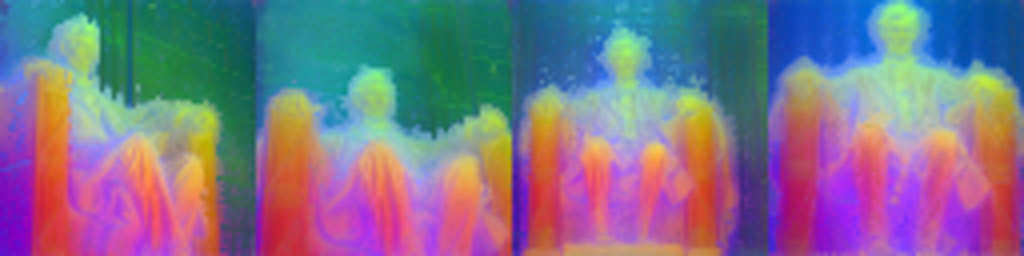} \\
[0mm]

% ===== Layer 8 =====
\rotatebox{90}{\small Layer 16} &
\includegraphics[width=\linewidth]{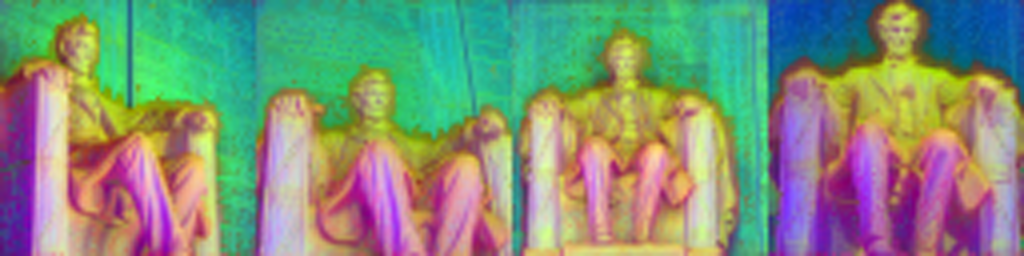} \\
[0mm]

% ===== Layer 10 =====
\rotatebox{90}{\small Layer 20} &
\includegraphics[width=\linewidth]{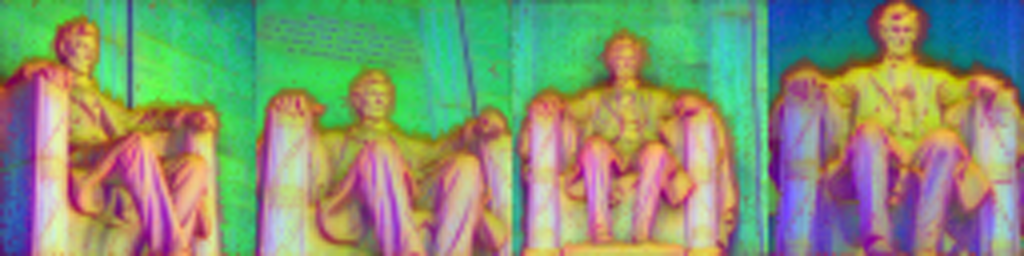} \\
[0mm]

% ===== Layer 12 =====
\rotatebox{90}{\small Layer 24} &
\includegraphics[width=\linewidth]{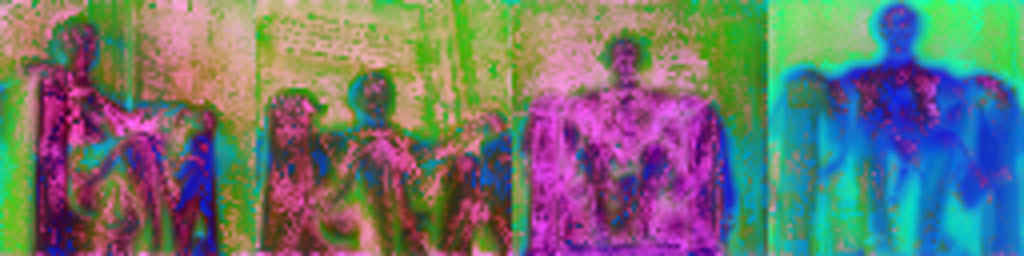} \\

\vspace{3mm}

% ===== DINOv3 =====
\rotatebox{90}{\small DINOv3} &
\includegraphics[width=\linewidth]{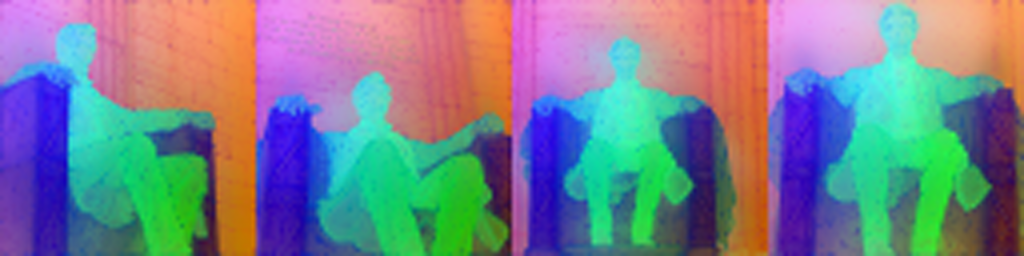} \\

\end{tabular}

\caption{PCA feature visualization across layers of \themethod-L. DINOv3 features are plotted on the last line.}
\label{fig:per_layer_fmap}
\end{figure}
\begin{figure*}[h]
    \centering % 居中整个 figure
    \begin{subfigure}[b]{0.16\linewidth}
        \includegraphics[width=\linewidth]{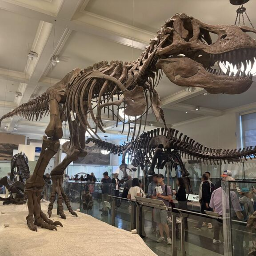}
    \end{subfigure}
    \hfill
    \begin{subfigure}[b]{0.16\linewidth}
        \includegraphics[width=\linewidth]{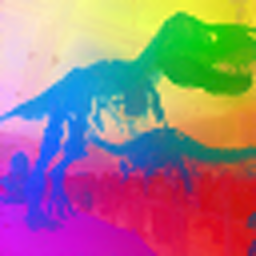}
    \end{subfigure}
    \hfill
    \begin{subfigure}[b]{0.16\linewidth}
        \includegraphics[width=\linewidth]{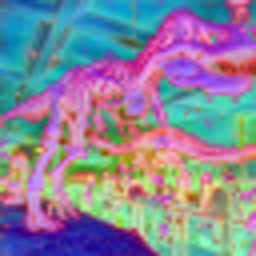}
    \end{subfigure}
    \hspace{0.01\linewidth}
    \begin{subfigure}[b]{0.16\linewidth}
        \includegraphics[width=\linewidth]{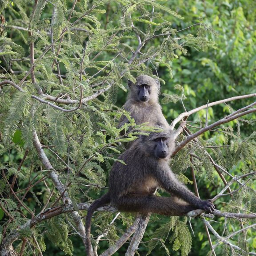}
    \end{subfigure}
    \hfill
    \begin{subfigure}[b]{0.16\linewidth}
        \includegraphics[width=\linewidth]{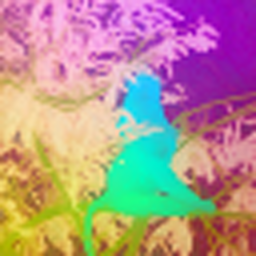}
    \end{subfigure}
    \hfill
    \begin{subfigure}[b]{0.16\linewidth}
        \includegraphics[width=\linewidth]{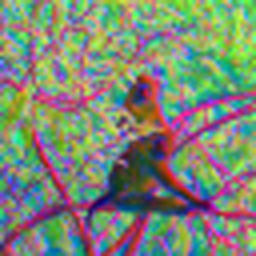}
    \end{subfigure}

    \begin{subfigure}[b]{0.16\linewidth}
        \includegraphics[width=\linewidth]{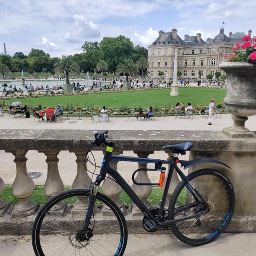}
    \end{subfigure}
    \hfill
    \begin{subfigure}[b]{0.16\linewidth}
        \includegraphics[width=\linewidth]{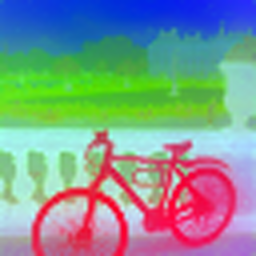}
    \end{subfigure}
    \hfill
    \begin{subfigure}[b]{0.16\linewidth}
        \includegraphics[width=\linewidth]{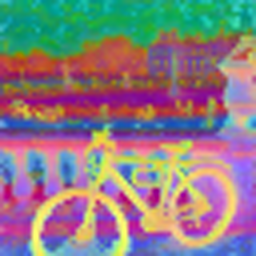}
    \end{subfigure}
    \hspace{0.01\linewidth}
    \begin{subfigure}[b]{0.16\linewidth}
        \includegraphics[width=\linewidth]{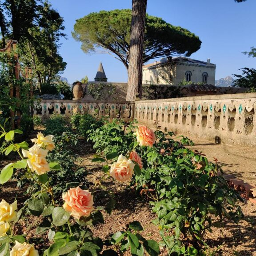}
    \end{subfigure}
    \hfill
    \begin{subfigure}[b]{0.16\linewidth}
        \includegraphics[width=\linewidth]{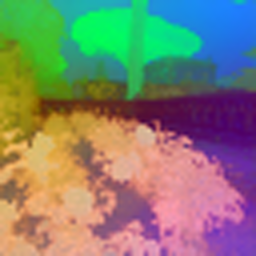}
    \end{subfigure}
    \hfill
    \begin{subfigure}[b]{0.16\linewidth}
        \includegraphics[width=\linewidth]{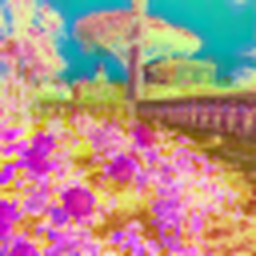}
    \end{subfigure}

    \begin{subfigure}[b]{0.16\linewidth}
        \includegraphics[width=\linewidth]{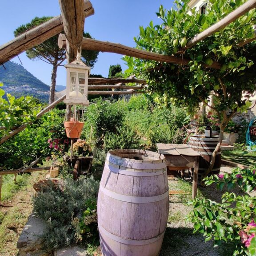}
        \caption*{Input}
    \end{subfigure}
    \hfill
    \begin{subfigure}[b]{0.16\linewidth}
        \includegraphics[width=\linewidth]{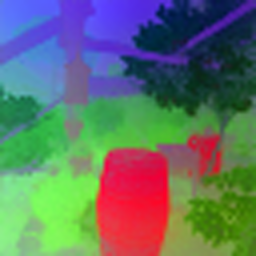}
        \caption*{DINOv3}
    \end{subfigure}
    \hfill
    \begin{subfigure}[b]{0.16\linewidth}
        \includegraphics[width=\linewidth]{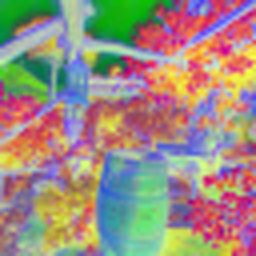}
        \caption*{\themethod}
    \end{subfigure}
    \hspace{0.01\linewidth}
    \begin{subfigure}[b]{0.16\linewidth}
        \includegraphics[width=\linewidth]{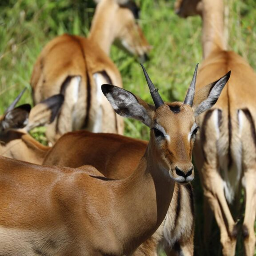}
        \caption*{Input}
    \end{subfigure}
    \hfill
    \begin{subfigure}[b]{0.16\linewidth}
        \includegraphics[width=\linewidth]{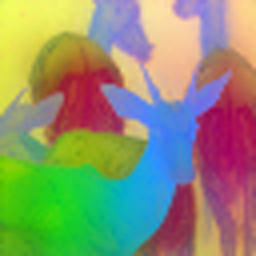}
        \caption*{DINOv3}
    \end{subfigure}
    \hfill
    \begin{subfigure}[b]{0.16\linewidth}
        \includegraphics[width=\linewidth]{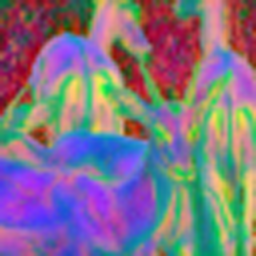}
        \caption*{\themethod}
    \end{subfigure}
    \caption{
        PCA feature visualization for single-view images, compared to DINOv3. % The features of layer 20 were selected to be displayed.
    }
    \label{fig:single_view_fmap}
\end{figure*}
\begin{figure*}[h]
    \centering % 居中整个 figure    
    \begin{subfigure}[b]{0.33\linewidth}
        \includegraphics[width=\linewidth]{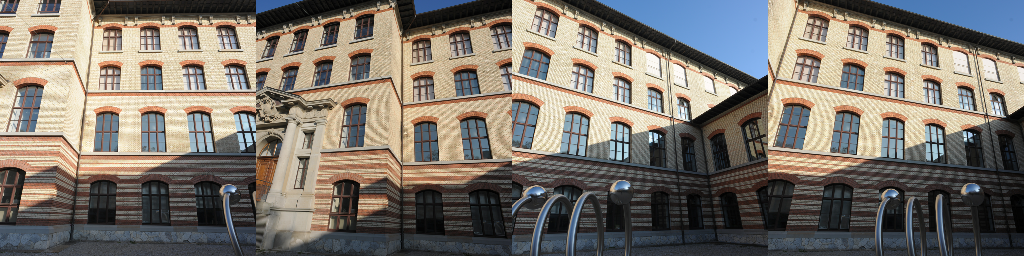}
    \end{subfigure}
    \hfill
    \begin{subfigure}[b]{0.33\linewidth}
        \includegraphics[width=\linewidth]{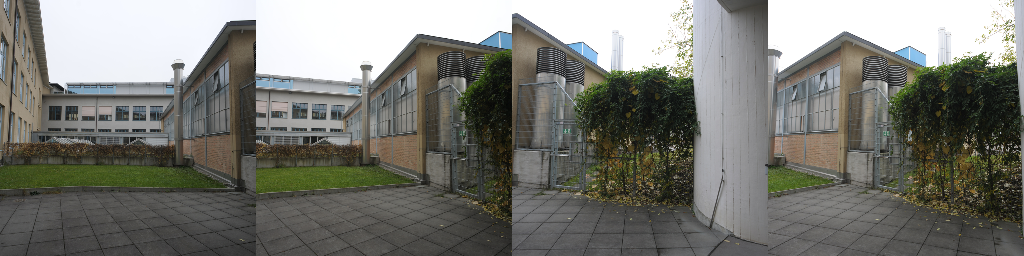}
    \end{subfigure}
    \hfill
    \begin{subfigure}[b]{0.33\linewidth}
        \includegraphics[width=\linewidth]{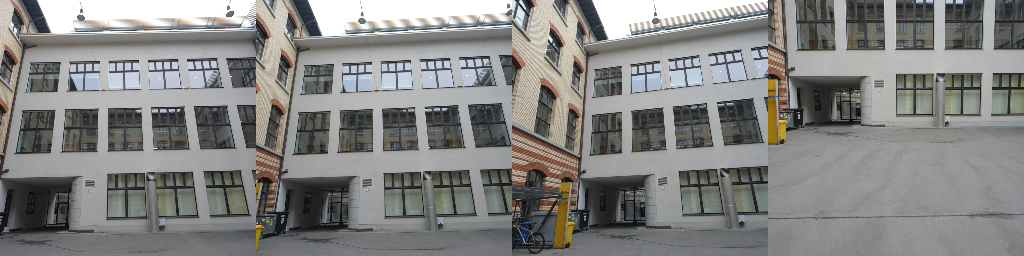}
    \end{subfigure}

    \begin{subfigure}[b]{0.33\linewidth}
        \includegraphics[width=\linewidth]{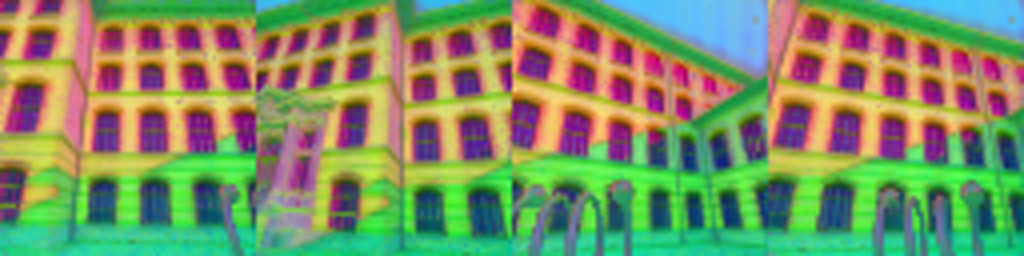}
    \end{subfigure}
    \hfill
    \begin{subfigure}[b]{0.33\linewidth}
        \includegraphics[width=\linewidth]{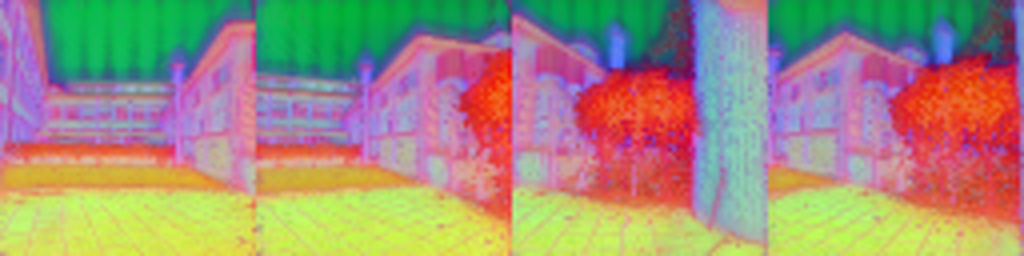}
    \end{subfigure}
    \hfill
    \begin{subfigure}[b]{0.33\linewidth}
        \includegraphics[width=\linewidth]{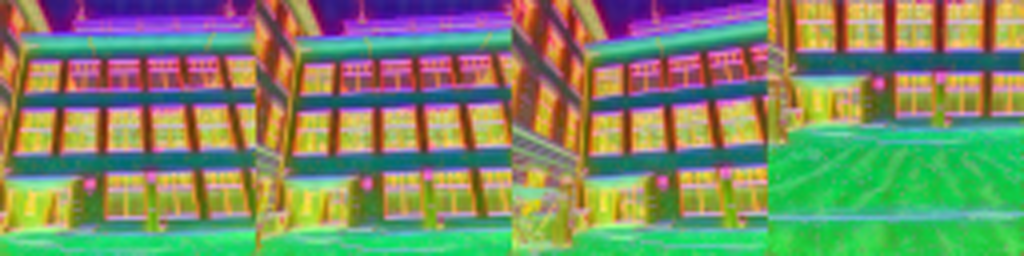}
    \end{subfigure}

    \vspace{1mm}

    \begin{subfigure}[b]{0.33\linewidth}
        \includegraphics[width=\linewidth]{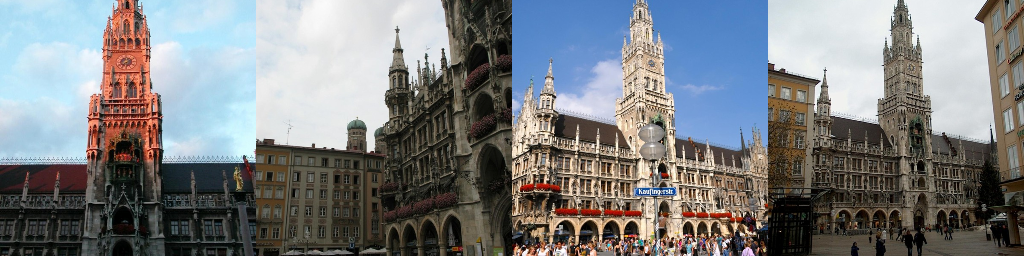}
    \end{subfigure}
    \hfill
    \begin{subfigure}[b]{0.33\linewidth}
        \includegraphics[width=\linewidth]{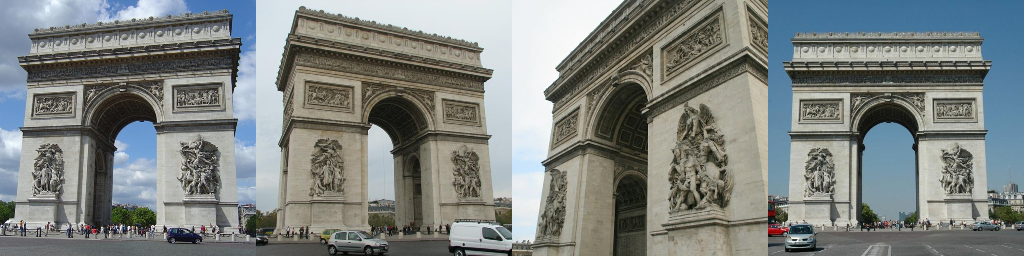}
    \end{subfigure}
    \hfill
    \begin{subfigure}[b]{0.33\linewidth}
        \includegraphics[width=\linewidth]{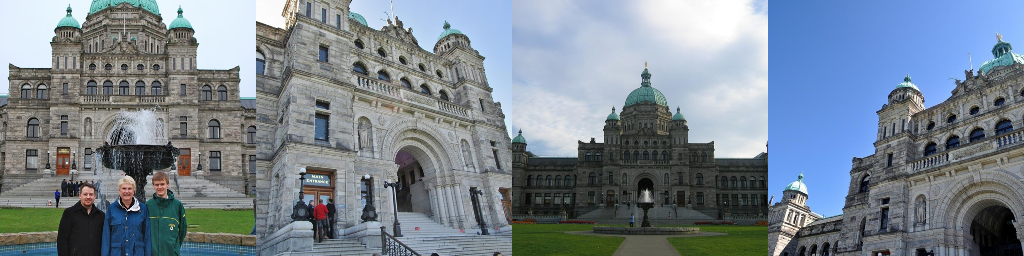}
    \end{subfigure}

    \begin{subfigure}[b]{0.33\linewidth}
        \includegraphics[width=\linewidth]{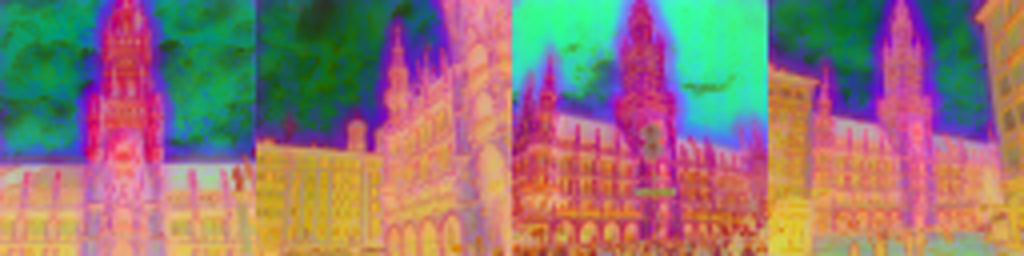}
    \end{subfigure}
    \hfill
    \begin{subfigure}[b]{0.33\linewidth}
        \includegraphics[width=\linewidth]{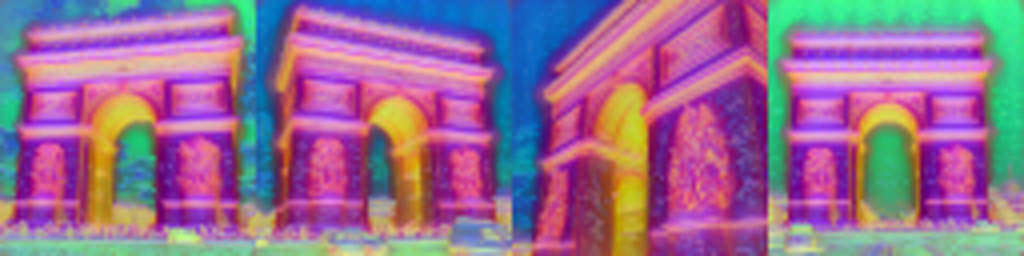}
    \end{subfigure}
    \hfill
    \begin{subfigure}[b]{0.33\linewidth}
        \includegraphics[width=\linewidth]{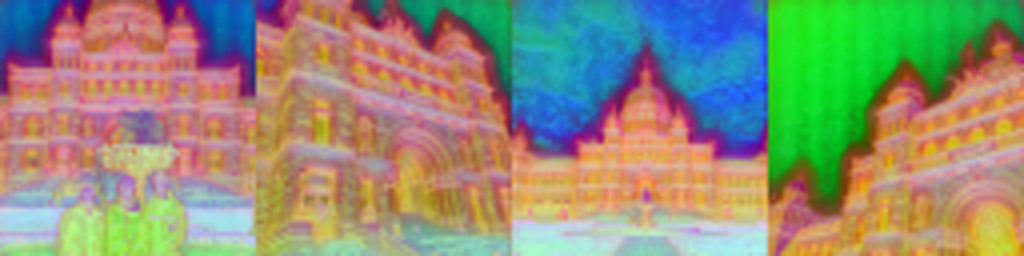}
    \end{subfigure}
    \caption{
        PCA feature visualization for multi-view images. % The features of layer 20 were selected to be displayed.
    }
    \label{fig:multi_view_fmap}
\end{figure*}

\subsection{Qualitative Feature Analysis}

In this section, we qualitatively analyze \themethod's dense feature representations. To this end, we project the high-dimensional feature space into three dimensions using principal component analysis (PCA), and visualize the resulting 3D embeddings by mapping them to RGB. All experiments use \themethod-L as the feature extractor. \cref{fig:per_layer_fmap} shows feature maps from layers 4, 8, 12, 16, 20, and 24. Notably, the 20th layer yields the most visually interpretable representation. In contrast, the final layer exhibits degraded visual structure due to the pixel-level reconstruction objective: its supervision signal is low-dimensional and highly ambiguous, which corrupts the high-level semantics in the features. We use the 20th layer for all subsequent visualizations.

\cref{fig:single_view_fmap} compares Muskie's single-view feature maps with those produced by DINOv3. While DINOv3 features display strong semantic consistency, they lose fine-grained structural details. This property benefits high-level tasks that rely on semantic abstraction but is less suitable for low-level reconstruction tasks that depend on precise feature matching. In contrast, Muskie's features preserve detailed geometry and object boundaries and exhibit strong cross-view consistency, making them substantially more suitable for reconstruction-oriented downstream applications. Additional multi-view visualizations are provided in \cref{fig:multi_view_fmap}, where corresponding points across views show consistent features that are largely invariant to illumination and viewpoint changes.

\subsection{Computational complexity}

As shown in \cref{tab:runtime_comparison}, we evaluate the inference runtime of our proposed Muskie backbone against the state-of-the-art frame-wise baseline, DINOv3.
Measurements are conducted using a single {NVIDIA A100 GPU} with {bfloat16} precision. 
Input images have a resolution of $224 \times 224$.
Both models leverage PyTorch's optimized \texttt{scaled\_dot\_product\_attention} operator for fair comparison.
While DINOv3 exhibits linear scaling ($\mathcal{O}(N)$) due to independent frame processing, Muskie employs global attention for half layers, theoretically leading to quadratic scaling ($\mathcal{O}(N^2)$). 
This sacrifice in inference speed is a necessary trade-off to move beyond the limitations of frame-wise procesing.

\newpage
{
    \small
    \bibliographystyle{ieeenat_fullname}
    \bibliography{refs_my,refs,vedaldi_general,vedaldi_specific}
}
\end{document}